%% file: neurips_2026.tex
\documentclass{article}

\usepackage[preprint]{neurips_2026}

\usepackage[utf8]{inputenc} 
\usepackage[T1]{fontenc}    
\usepackage{hyperref}       
\usepackage{url}            
\usepackage{booktabs}       
\usepackage{amsfonts}       
\usepackage{nicefrac}       
\usepackage{microtype}      
\usepackage{xcolor}         

\usepackage{microtype}
\usepackage{graphicx}
\usepackage{subcaption}
\usepackage{booktabs} 
\usepackage{amsmath}
\usepackage{amssymb}
\usepackage{mathtools}
\usepackage{amsthm}
\usepackage{threeparttable}
\usepackage{hyperref}
\usepackage{url} 
\usepackage{booktabs}
\usepackage{multirow}
\usepackage{graphicx}
\usepackage{makecell}
\usepackage{subcaption}
\usepackage{caption}
\usepackage{subcaption}
\usepackage{placeins} 

\title{Vision Hopfield Memory Networks for\\ Image Recognition}

\author{%
\begin{tabular}{c}
\textbf{Jianfeng Wang}\textsuperscript{1}\thanks{Equal contribution.}\hspace{0.2em} 
\thanks{Corresponding author: jianfengwang1991@gmail.com. Work initiated at University of Oxford. Part of this work was completed after the author's departure.}
\hspace{1.2em}
\textbf{Amine M'Charrak}\textsuperscript{1}\footnotemark[1]
\hspace{1.2em}
\textbf{Luk Koska}\textsuperscript{2}
\hspace{1.2em}
\textbf{Xiangtao Wang}\textsuperscript{2}
\\[0.35em]
\textbf{Daniel Petriceanu}\textsuperscript{2}
\hspace{1.2em}
\textbf{Ruizhi Wang}\textsuperscript{2}
\hspace{1.2em}
\textbf{Michael Bumbar}\textsuperscript{2}
\hspace{1.2em}
\textbf{Luca Pinchetti}\textsuperscript{1} 
\\[0.35em]
\textbf{Thomas Lukasiewicz}\textsuperscript{1,\ 2}
\\[0.5em]
{\normalfont\textsuperscript{1}Department of Computer Science, University of Oxford, United Kingdom}
\\[0.15em]
{\normalfont\textsuperscript{2}Faculty of Informatics, Vienna University of Technology, Austria}
\end{tabular}%
}

\begin{document}

\maketitle

\begin{abstract}

Recent vision backbones, such as Transformer families and state-space models like Mamba, have achieved remarkable progress on image recognition. Despite their empirical success, these architectures remain far from the computational principles of the human brain, often demanding enormous amounts of training data while offering limited interpretability. In this work, we propose the Vision Hopfield Memory Network (V-HMN), a brain-inspired vision backbone that integrates hierarchical memory mechanisms across layers with iterative refinement updates. 
Specifically, V-HMN incorporates local Hopfield modules that provide associative memory dynamics at the image patch level, global Hopfield modules that function as episodic memory for contextual modulation, and a predictive-coding-inspired refinement rule for iterative error correction. By organizing these memory-based modules hierarchically, V-HMN captures both local and global dynamics in a unified framework.  
Memory retrieval exposes the relationship between inputs and stored patterns, providing a prototype-based form of interpretability through explicit memory retrieval, while the reuse of stored patterns improves data efficiency. 
This brain-inspired design therefore enhances data efficiency and provides a prototype-based form of interpretability compared to existing self-attention- or state-space-based approaches. 
We conducted extensive experiments on public image classification benchmarks. V-HMN achieves strong performance on small- and medium-scale benchmarks, and remains competitive with widely adopted backbone architectures on ImageNet despite minimal architectural tuning, while offering improved data efficiency and a prototype-based form of interpretability. These findings highlight the potential of V-HMN as a memory-centric alternative to standard vision backbones, thereby bridging brain-inspired computation with modern machine learning.

\end{abstract}

\section{Introduction}
Vision backbones have experienced significant changes in recent years. 
Starting with AlexNet \citep{krizhevsky2012alexnet} and its revolutionary performance, convolutional neural networks (CNNs) attracted the attention of researchers, leading to the design of advanced architectures such as VGG \citep{simonyan2015vgg} and ResNet \citep{he2016resnet}. 
Subsequently, the evolution of network architectures in natural language processing, particularly the Transformer \citep{vaswani2017attention}, gave rise to its vision counterpart, the Vision Transformer (ViT) \citep{dosovitskiy2020vit}, which achieved promising results on computer vision benchmarks. 
Building on this trend, a variety of alternative architectures have been proposed, such as MLP-Mixer \citep{tolstikhin2021mlpmixer}, MetaFormer \citep{yu2022metaformer}, and more recently, state-space models like Vision Mamba (Vim) \citep{zhu2024visionmamba}, further enriching the landscape of vision backbones.

However, these models do not fundamentally address some of the long-standing challenges in deep learning. 
Specifically, they are not data-efficient and usually require large-scale datasets for training. 
Moreover, they lack biological plausibility, as their learning mechanisms differ substantially from how the human brain operates. 
In terms of data efficiency, current models rely heavily on extensive supervised training and large annotated datasets, which limit their applicability in domains where data collection is time-consuming or even infeasible. 
In contrast, humans are able to learn robust concepts from very limited examples, pinpointing the gap between artificial and natural learning. 
As for biological plausibility, deep learning architectures and optimization methods are largely engineered for computational convenience rather than grounded in neuroscience. 
For example, the conventional feedforward architecture overlooks key properties of the human brain, such as associative memory retrieval \citep{ramsauer2021hopfield} and predictive error correction \citep{rao1999predictive,friston2005theory}.

To deal with these challenges, we propose a new vision model, named \textbf{Vision Hopfield Memory Network (V-HMN)}. 
V-HMN departs from conventional feedforward or self-attention-only designs by making \emph{content-addressable associative memory}\footnote{%
In Hopfield networks, content-addressable memory refers to the ability to retrieve a stored pattern by filling in missing or noisy parts of the input.}  a central computational primitive in each block.
Concretely, V-HMN employs two complementary memory paths: 
(i) a \emph{local window memory} that collects $k \!\times\! k$ neighborhoods and performs Hopfield-style retrieval to denoise and complete local patterns; and 
(ii)~a~\emph{global template path} that forms a scene-level query via global pooling, retrieves a global prototype from memory, and injects it back into all tokens as a context prior. 
Both memory paths update features through an iterative refinement step with a learnable strength parameter. 
This mechanism can be viewed as a lightweight form of predictive-coding dynamics, where representations are gradually corrected toward memory-predicted patterns. 
In this way, the network gains an error-corrective feedback process that is absent in conventional feedforward models.

\section{Related Work}
We now briefly review related works, including existing vision backbones, associative~memory and modern Hopfield networks, and brain-inspired predictive coding frameworks.

\subsection{Vision backbones}
Early advances in vision backbones were driven by convolutional neural networks (CNNs), such as AlexNet, VGG, and ResNet. 
While these models achieved remarkable progress, recent research has shifted toward alternative token-mixing paradigms. 
ViT showed that a pure Transformer on image patches can rival CNNs at scale \citep{dosovitskiy2020vit}, and hierarchical designs like Swin Transformer (Swin-ViT) \citep{liu2021swin} improved efficiency through shifted local windows. 
In addition, a line of hybrid architectures attempts to combine the complementary strengths of CNNs and Transformers. 
Representative examples include ConViT \citep{d2021convit}, which introduces soft convolutional inductive biases into attention layers, and CoaT \citep{xu2021coat}, which integrates co-scale convolution with multi-head attention for better local-global trade-offs. 
Such hybrids highlight the ongoing interest in balancing locality and global context within a unified backbone. 
Beyond attention, MLP-based models (e.g., MLP-Mixer \citep{tolstikhin2021mlpmixer}) and MetaFormer\footnote{Throughout our experiments, 
we follow common practice and instantiate MetaFormer using PoolFormer, 
which is the default implementation adopted in prior work.} frameworks \citep{yu2021metaformer,yu2022metaformer} demonstrated that different mixers can operate within a similar architectural scaffold. 
Most recently, state-space models (SSMs) have emerged as competitive backbones. 
S4 introduced structured SSMs for long sequences \citep{gu2021s4}, and Mamba extended this idea with selective input-dependent dynamics \citep{gu2023mamba}. 
Vision-specific variants such as Vim \citep{zhu2024visionmamba} and VMamba \citep{vmamba2024} show promising results with linear-time complexity. 
While these advances broaden the landscape of vision backbones, they typically do not incorporate explicit, content-addressable memory mechanisms.

A recent attempt to address this limitation is the Associative Transformer (AiT) \citep{sun2025associative}, which introduces a global workspace layer where memory slots are written via bottleneck attention and retrieved through Hopfield-style dynamics to refine token embeddings. 
While AiT demonstrates the potential of integrating associative memory into Transformer architectures, it remains primarily Transformer-based, where memory acts as an auxiliary component and self-attention still plays a central role in token interactions. 
By contrast, V-HMN is a memory-centric backbone in which local and global Hopfield modules serve as the primary mechanism for token mixing, largely replacing self-attention. 
This design makes V-HMN conceptually simpler and offers improved interpretability and data efficiency, positioning memory not as an add-on but as the core of the backbone itself.

\subsection{Associative memory and modern Hopfield networks}
Modern Hopfield Networks (MHNs) revisit content-addressable memory with continuous states and an energy function that yields exponentially large storage and single-step retrieval in theory \citep{ramsauer2021hopfield}. This line has been integrated into practical deep architectures via a differentiable Hopfield layer and applied beyond vision (e.g., retrieval, pooling, representation learning) \citep{ramsauer2021hopfield, fuerst2021cloob}. Recent works refine robustness and capacity, and study retrieval dynamics under modern settings \citep{wu2024uniform, hu2024outlierefficient}. Compared to self-attention, Hopfield-style modules maintain a \emph{persistent} memory bank with explicit slots (prototypes), enabling interpretable slot activations and prototype-token alignments. Our V-HMN leverages this by combining a local window memory (prototype completion/denoising) and a global template path (scene-level prior), both trained end-to-end.

Beyond Hopfield-style associative retrieval, a broader line of work has explored \emph{prototype memories} and \emph{external memory banks} as mechanisms for improving data efficiency. Early metric-based few-shot learning methods such as matching networks \citep{vinyals2016matching} and prototypical networks \citep{snell2017prototypical} explicitly store class-level prototypes in an embedding space and perform inference via metric retrieval, showing that maintaining persistent prototypes can substantially improve generalization under limited supervision. Subsequent extensions, including relation networks \citep{sung2018relation} and MetaOptNet \citep{lee2019metaoptnet}, further refine prototype-based retrieval by learning similarity functions or optimizing embedding geometry for more reliable few-shot generalization. 
Prototype memory has also been explored in generative modeling.  
The approach of \citep{li2022prototype} maintains a learned bank of visual prototypes and retrieves 
them via attention to guide synthesis from only a handful of examples. Similarly, 
\citep{li2022prototype} introduce a prototype-conditioned generative mechanism in which retrieved 
prototypes act as structural priors that stabilize low-data generation. Both lines of work demonstrate 
that reusing persistent prototypes can effectively expand the statistical support of limited datasets 
and improve sample efficiency in generative scenarios.

\subsection{Predictive-coding-inspired iterative refinement}
Predictive coding (PC) is a long-standing theory in neuroscience that frames perception as iterative error minimization between predictions and sensory input \citep{rao1999predictive, friston2005theory}. In computational neuroscience, PC networks (PCNs) have been studied as a biologically plausible alternative to backpropagation \citep{whittington2019theories}, while early deep learning variants such as PredNet applied the idea to video prediction with hierarchical recurrent modules \citep{lotter2017prednet}. More recent works have attempted to formalize PCNs in machine learning, drawing connections to variational inference, energy-based models, and equilibrium propagation \citep{millidge2022pcfuture, vzwol2024pcn}. Despite their theoretical appeal, PC-inspired models remain limited to small-scale or domain-specific settings, in part due to optimization difficulties and inefficiency in large-scale vision tasks. In this work, we do not implement full PC inference; instead, our blocks perform a lightweight, learnable refinement toward memory-predicted prototypes. This provides an interpretable, error-corrective step that is inspired by PC principles, connecting HMN-style memory refinement with a brain-inspired narrative, while keeping the backbone simple and scalable.

\section{Methodology}
In this section, we give an overview of the overall architecture of V-HMN, outlining how images are processed from patch embeddings to memory-based refinement blocks and finally to classification.

\subsection{Overall Architecture}
The  V-HMN is designed as a memory-centric vision backbone. 
An input image is first projected into a sequence of image patch tokens. 
These tokens are then processed by a stack of HMN blocks, each integrating local and global Hopfield memory modules\footnote{Appendix~\ref{appendix:hierarchy-invariances} provides a qualitative discussion clarifying which invariances arise from the hierarchical local-global memory design and which are attributable to standard components such as augmentation and pooling.}. 
Finally, the sequence is  aggregated by attention pooling, and fed into a linear classifier. 
In contrast to convolutional backbones (ResNet \citep{he2016resnet}), self-attention-based designs (ViT \citep{dosovitskiy2020vit}), or state-space models (Vim \citep{zhu2024visionmamba}), V-HMN replaces the underlying token-mixing operation with explicit associative memory retrieval. 
This makes memory refinement (not convolution, self-attention, or recurrence) the core building block of the backbone.

\subsection{Local and Global Hopfield Memory Modules}
\label{Sec32}
Each HMN block contains two complementary modules that together capture fine-grained local structure and holistic global context.

\textbf{Local memory.}  
For each token, its $k \!\times\! k$ spatial neighborhood is unfolded with padding, flattened, and linearly projected to form a latent query for that token. Hopfield retrieval is then performed against a class-balanced memory bank that stores prototype features. 
The retrieved prototype refines the local representation by stabilizing noisy features and completing partial patterns. 
The initial and refined representations are concatenated, projected back to the embedding dimension, and subsequently fused with the global branch before being added to the input representation via a skip connection.
This mechanism parallels the role of convolutions or windowed attention, but operates through explicit prototype-based priors.

\begin{figure*}[t]
    \centering    
    \includegraphics[width=\linewidth]{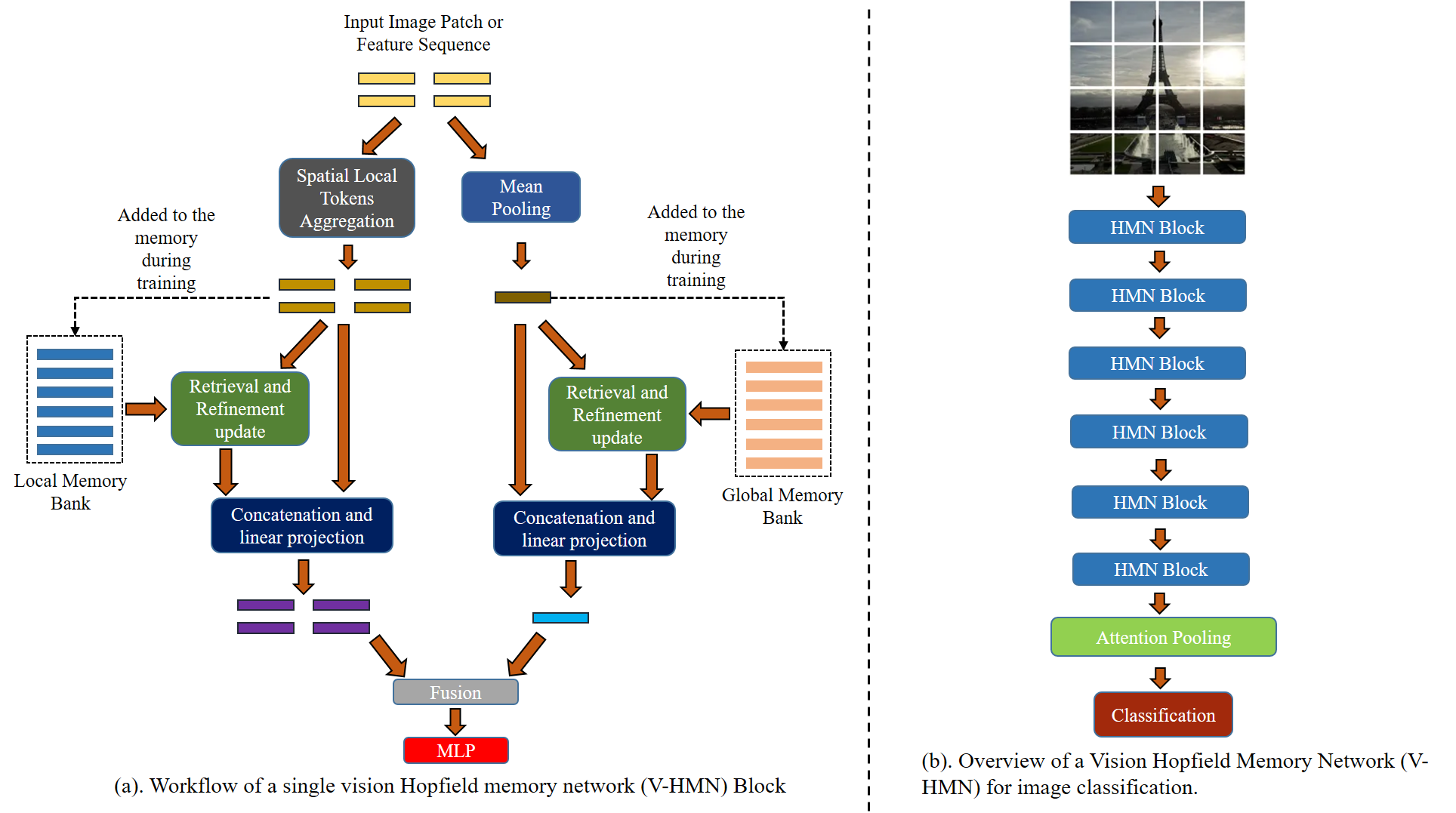}
    \caption{Overview of V-HMN. 
(a) Each HMN block refines features through local and global Hopfield memory retrieval, rather than convolution or self-attention. 
(b) A deep backbone is constructed by stacking HMN blocks, with attention pooling and a linear head for image classification.}
 
    \label{fig:overview} 
\end{figure*}

\textbf{Global memory.} 
To provide scene-level context, the global branch first mean-pools all tokens to form a query vector. 
This query interacts with a global memory bank through Hopfield retrieval, producing a prototype that captures global semantics. 
The result is broadcast to all tokens and integrated with the local path.

The overall workflow of a V-HMN block and its integration into the classification backbone are illustrated in Figure~\ref{fig:overview}.  
Figure~\ref{fig:overview}~(a) details a single block: the input token sequence first branches into 
(i) a \emph{local memory} path that aggregates $k\times k$ spatial neighborhoods, performs associative retrieval with iterative refinement, and concatenates the refined and initial representations; and (ii) a \emph{global memory} path that mean-pools all tokens to form a scene-level query, applies Hopfield-based retrieval and iterative refinement, and concatenates the refined and initial queries. 
The two paths are then fused by summation, and the fused representation is passed through a lightweight two-layer MLP, forming the output of the MHN block.
In particular, the associative memory retrieval can be formalized as follows. 
Given a representation $z \in \mathbb{R}^{D}$ and a memory bank 
$M \in \mathbb{R}^{K\times D}$ with $K$ prototype slots, we first 
$\ell_2$-normalize both $z$ and the memory slots to obtain cosine similarities: 
\begin{equation}
\hat z = \frac{z}{\|z\|_2}, \qquad
\hat M_j = \frac{M_j}{\|M_j\|_2}, \; j=1,\dots,K .
\label{eq:normalization}
\end{equation}
Retrieval weights and the retrieved prototype are then computed by 
\begin{equation}
\alpha = \mathrm{softmax}\!\big(\sqrt{D}\,\hat z\,\hat M^{\top}\big)
\in \mathbb{R}^{K},
\qquad
m = \sum\nolimits_{j=1}^{K} \alpha_j\, M_j
\in \mathbb{R}^{D}.
\label{eq:retrieval}
\end{equation}
 where $M_j$ is the $j$-th \emph{unnormalized} memory slot, 
$\alpha$ are normalized weights ($\sum_j \alpha_j=1$), and $m$ is the retrieved prototype. 
The additional scaling factor $\sqrt{D}$ is introduced, because cosine similarities 
$\hat z \hat M^\top_j$ have variance in the order of $1/D$: multiplying by $\sqrt{D}$ 
restores the logits to approximately unit variance \footnote{A detailed proof is provided in Appendix~\ref{appendix:variance-proof}.}, preventing the softmax distribution 
from becoming overly flat and yielding sharper, more discriminative retrieval weights.

Figure~\ref{fig:overview}~(b) shows the full classification backbone formed by stacking multiple HMN blocks in depth, followed by attention pooling and a linear classifier. Specifically, attention pooling performs a weighted combination over all tokens to produce a single representation, which can be defined as:
\begin{equation}
\alpha = \mathrm{softmax}\!\left(H\,W_{\text{att}}\right)
\in \mathbb{R}^{N},
\qquad
v = \sum\nolimits_{i=1}^{N} \alpha_i\, H_i
\in \mathbb{R}^{D}.
\label{eq:attention_pooling}
\end{equation}
where $H \in \mathbb{R}^{N\times D}$ denotes the token representations after the final block, 
$W_{\text{att}}\in\mathbb{R}^{D\times 1}$ is a learnable scoring vector that assigns importance weights to tokens, 
$\alpha$ are the normalized attention weights ($\sum_i \alpha_i=1$), 
and $v$ is the pooled representation fed into the classifier. 

During training, both local and global modules maintain their own class-balanced memory banks. 
Unlike parametric weights, these banks are explicitly written with real sample embeddings at each block: 
the local bank stores projected patch-neighborhood features, while the global bank stores pooled scene-level features. 
Each bank is organized as a per-class ring buffer with fixed capacity, ensuring that all classes are allocated equal slots. 
As training proceeds, new embeddings replace the oldest ones within each class, yielding a continually refreshed and balanced set of prototypes. 
During inference, the banks are frozen and no longer updated, so retrieval always operates on stable prototypes that persist across tasks.

\subsection{Iterative Refinement}
The central operation in both local and global modules is an iterative refinement rule. 
Given a current representation $z$ and a retrieved prototype $m$, the update is
\begin{equation}
z^{(t+1)} = z^{(t)} + \beta \,(m - z^{(t)}),
\label{iteration}
\end{equation}
where $\beta$ is a learnable update strength and $t$ denotes the refinement step. 
This mechanism can be viewed as a \emph{predictive-coding-inspired} update 
\footnote{A more detailed discussion of the connection between our refinement rule and predictive-coding (PC) dynamics is provided in Appendix~\ref{appendix:refinement-pc}. 
Appendix~\ref{appendix:bio-plausibility} further discusses the biological plausibility and cortical inspiration of the overall V-HMN design.}: the prototype $m$ provides a memory-based prediction, while the residual $(m - z^{(t)})$ acts as a prediction error that gradually corrects the current representation, in line with the associative memory’s role of filling in missing or noisy information.
In contrast to full predictive coding networks that maintain explicit error units and multi-layer recurrent inference, V-HMN adopts a lightweight refinement loop where only a few steps are sufficient in practice. This refinement design leverages \emph{persistent, content-addressable prototypes} that are shared across samples, 
and it yields two key benefits: (i) improved data efficiency, as stored prototypes provide reusable priors; and (ii) enhanced interpretability, since prototype activations directly expose the memory patterns supporting each decision.

\section{Experiments}

We now report on our experiments on five public image classification benchmarks, including  CIFAR-10 \citep{krizhevsky2009learning}, CIFAR-100 \citep{krizhevsky2009learning}, SVHN \citep{netzer2011reading}, Fashion-MNIST \citep{xiao2017fashion}, and ImageNet-1k \citep{deng2009imagenet}. To save space, implementation details are given in the appendix.


\subsection{Ablation Studies}

\begin{table}[t]
\centering
\caption{Data efficiency of V\mbox{-}HMN on CIFAR-10, CIFAR-100, and Fashion-MNIST with different fractions of labeled training samples. Reported values are top-1 test accuracy (\%).}
\vspace{1ex}
\label{tab:data-fractions}
\resizebox{0.7\linewidth}{!}{\begin{tabular}{c|c|c|c}
\toprule
\textbf{Fraction of training data} & \textbf{CIFAR-10} & \textbf{CIFAR-100} & \textbf{Fashion-MNIST} \\
\midrule
10\%  &  $80.22_{\pm 0.29 }$  & $43.21_{\pm 1.07 }$ & $89.18_{\pm 0.16 }$ \\
30\%  &   $88.67_{\pm 0.21 }$ & $62.42_{\pm 0.29 }$   & $91.04_{\pm 0.22  }$  \\
50\%  &  $91.19_{\pm  0.38 }$ &  $68.93_{\pm  0.35 }$  &  $91.53_{\pm 0.12}$ \\
\bottomrule
\end{tabular}
}
\end{table}

\begin{table}[t]
\centering
\caption{Comparison of data efficiency across models with 10\% and 30\% labeled training data on CIFAR-10, CIFAR-100, and Fashion-MNIST. 
Top-1 test accuracy (\%) are reported as mean $\pm$ standard deviation over 3 seeds. 
Baselines include ViT \citep{dosovitskiy2020vit}, 
Swin-ViT \citep{liu2021swin}, 
MLP-Mixer \citep{tolstikhin2021mlpmixer}, 
MetaFormer \citep{yu2022metaformer}, 
Vim \citep{zhu2024visionmamba}, 
and AiT \citep{sun2025associative}.}
\vspace{1ex}
\label{tab:data-efficiency-all-models}

\resizebox{0.95\linewidth}{!}{
\begin{tabular}{l|cc|cc|cc}
\toprule
\multirow{2}{*}{\textbf{Model}} 
& \multicolumn{2}{c|}{\textbf{CIFAR-10}}
& \multicolumn{2}{c|}{\textbf{CIFAR-100}}
& \multicolumn{2}{c}{\textbf{Fashion-MNIST}} \\

\cmidrule(lr){2-3}
\cmidrule(lr){4-5}
\cmidrule(lr){6-7}

& 10\% & 30\%
& 10\% & 30\%
& 10\% & 30\% \\
\midrule

ViT
& $72.73_{\pm 0.42}$
& $83.94_{\pm 0.33}$
& $40.48_{\pm 0.80}$
& $57.40_{\pm 0.79}$
& $87.17_{\pm 0.40}$
& $89.71_{\pm 0.45}$ \\

Swin-ViT
& $70.37_{\pm 0.94}$
& $80.89_{\pm 0.24}$
& $35.25_{\pm 0.59}$
& $51.77_{\pm 0.60}$
& $88.42_{\pm 0.48}$
& $90.34_{\pm 0.15}$ \\

MLP-Mixer
& $76.14_{\pm 0.16}$
& $85.53_{\pm 0.40}$
& $41.94_{\pm 0.98}$
& $56.22_{\pm 0.75}$
& $87.16_{\pm 0.63}$
& $89.41_{\pm 0.13}$ \\

MetaFormer
& $49.92_{\pm 4.08}$
& $60.86_{\pm 6.05}$
& $19.39_{\pm 1.26}$
& $37.47_{\pm 3.19}$
& $85.02_{\pm 1.22}$
& $88.78_{\pm 1.03}$ \\

Vim
& $69.02_{\pm 2.13}$
& $79.58_{\pm 0.37}$
& $36.61_{\pm 1.25}$
& $49.67_{\pm 0.79}$
& $86.09_{\pm 1.03}$
& $88.18_{\pm 2.43}$ \\

AiT
& $67.89_{\pm 0.55}$
& $80.69_{\pm 1.36}$
& $35.84_{\pm 0.67}$
& $54.59_{\pm 0.87}$
& $84.33_{\pm 0.38}$
& $87.62_{\pm 0.32}$ \\

V-HMN (ours)
& $\boldsymbol{80.22_{\pm 0.29}}$
& $\boldsymbol{88.67_{\pm 0.21}}$
& $\boldsymbol{43.21_{\pm 1.07}}$
& $\boldsymbol{62.42_{\pm 0.29}}$
& $\boldsymbol{89.18_{\pm 0.16}}$
& $\boldsymbol{91.04_{\pm 0.22}}$ \\

\bottomrule
\end{tabular}
}
\end{table}

\paragraph{Data Efficiency.} 
We study the data efficiency of V\mbox{-}HMN under varying fractions of labeled training data. 
Table~\ref{tab:data-fractions} shows that accuracy improves steadily as the proportion of labeled data increases: even with only 10\% of the training set, V\mbox{-}HMN achieves competitive performance, while scaling to 30\% and 50\% further closes the gap to the full-data regime. 
This indicates that the associative memory mechanism provides strong inductive biases that reduce dependence on large-scale annotation. 

We further benchmark V\mbox{-}HMN against standard vision backbones under 10\% and 30\% labeled data (Table~\ref{tab:data-efficiency-all-models}). 
Across CIFAR-10, CIFAR-100, and Fashion-MNIST, V\mbox{-}HMN consistently outperforms widely used architectures such as ViT, Swin-ViT, MLP-Mixer, and MetaFormer, as well as more recent models like Vim. 
Importantly, V\mbox{-}HMN also surpasses AiT, which integrates memory slots into a Transformer backbone. 
This highlights the benefit of our design: rather than appending memory to an existing architecture, V\mbox{-}HMN makes associative memory the core computational primitive. 
The resulting prototype-based refinement provides stronger gains, especially in low-data settings, where stored prototypes act as reusable priors and compensate for scarce supervision\footnote{We provide an analysis in Appendix~\ref{appendix:data-efficiency-analysis} explaining why V-HMN exhibits data efficiency.}.

\begin{figure*}[t]
\centering
\begin{minipage}{0.5\textwidth}
    \centering
    \captionof{table}{Ablation study on the number of refinement iterations in V-HMN. Top-1 test accuracy (\%) are reported as mean $\pm$ standard deviation over 3 seeds.}
    \label{tab:iteration_ablation}
    
    \resizebox{0.95\textwidth}{!}{%
        \begin{tabular}{cccc}
        \toprule
        Iterations & CIFAR-10 & CIFAR-100 & Fashion-MNIST \\
        \midrule
        0 & $93.56_{\pm 0.10}$  & $75.84_{\pm 0.14}$ &  $92.05_{\pm 0.08}$ \\
        1 & $93.94_{\pm 0.11}$  & $76.58_{\pm 0.09}$  &  $92.27_{\pm 0.06}$ \\
        2 & $94.28_{\pm 0.13}$  & $76.59_{\pm 0.16}$  &  $92.48_{\pm 0.06}$\\
        3 & $93.99_{\pm 0.07}$  & $76.41_{\pm 0.09}$  &  $92.40_{\pm 0.05}$ \\
        \bottomrule
        \end{tabular}
    }
\end{minipage}
\hfill
\begin{minipage}{0.46\textwidth}
    \centering
    
    \captionof{figure}{Effect of refinement iteration on CIFAR-10 robustness across Gaussian noise, occlusion, and contrast corruptions.}
    \includegraphics[width=0.8\textwidth]{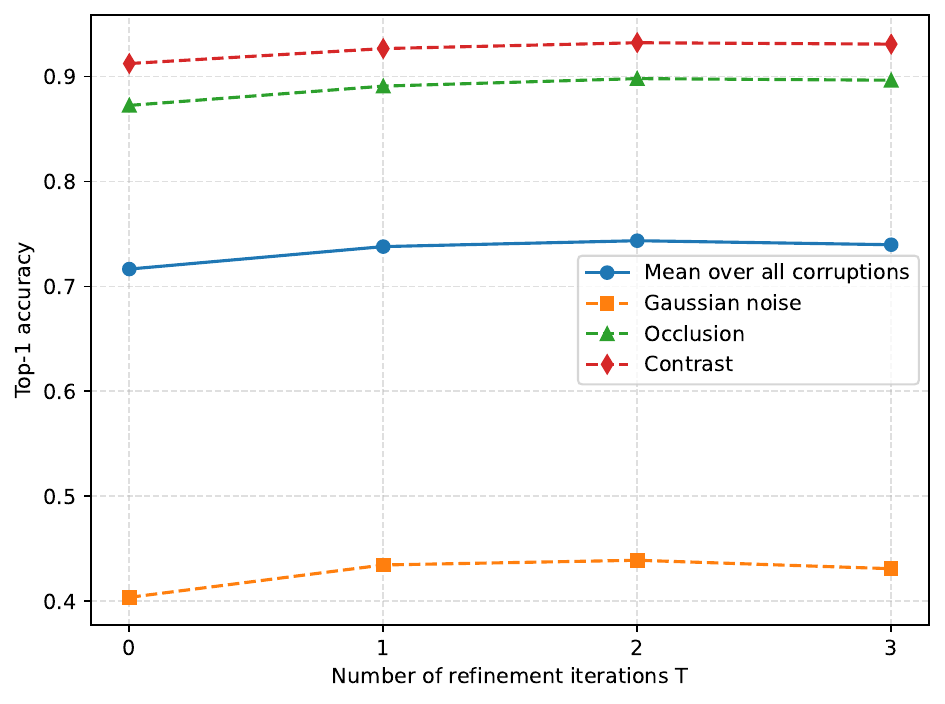}
    \label{fig:refinement_robustness}
\end{minipage} 
\end{figure*}

\paragraph{Iterative Refinement.}

Table~\ref{tab:iteration_ablation} summarizes the effect of varying the number of refinement iterations~$t$.
Here, $t=0$ disables the associative refinement loop entirely: the local and global branches still compute their feedforward projections, but no Hopfield retrieval or error-correction update (Eq.~\ref{iteration}) is applied. The memory banks remain allocated during training to keep the parameter count identical, but they are not read during inference.

Across all datasets, introducing even a single refinement step ($t=1$) yields consistent improvements (e.g., CIFAR-10: 93.56\% $\rightarrow$ 93.94\%; CIFAR-100: 75.84\% $\rightarrow$ 76.58\%). This confirms that the associative update provides meaningful benefits, although the underlying feedforward pathway is already strong. Performance peaks at $t=2$, while deeper unrolling offers no additional gain and can slightly degrade accuracy due to over-correction. This trend is consistent with predictive-coding models of cortical processing, where a small number of recurrent error-correction steps typically suffices to explain the input and further unrolling mainly increases computational cost~\citep{rao1999predictive,friston2005theory}.

In addition, we conduct robustness experiments to assess whether additional refinement iterations offer benefits beyond a single update. We evaluate CIFAR-10 models trained with different numbers of refinement steps under several corruptions: Gaussian noise (standard deviations 0.05, 0.10, 0.20, 0.30), random square occlusion (areas 0.05, 0.10, 0.20), and contrast scaling (factors 0.5, 0.75, 1.25, 1.5). As shown in Figure~\ref{fig:refinement_robustness}, we report both the mean top-1 accuracy across all corruptions and the per-corruption accuracy for each refinement depth. Averaged over all corruptions, accuracy increases from 71.65\% at $t=0$ to 73.79\% at $t=1$ and 74.35\% at $t=2$. The gains are most pronounced for occlusion and contrast: occlusion accuracy improves from 87.24\% ($t=0$) to 89.08\% ($t=1$) and 89.79\% ($t=2$), with similar improvements for contrast scaling. Overall, these results demonstrate that the predictive-coding-inspired refinement yields measurable and consistent robustness gains. 
Unless otherwise specified, we
fix the number of iterations to $t = 1$ in all other experiments for a balanced trade-off between
accuracy and efficiency.

Additional analyses of memory constructions, retrieval behavior, and runtime characteristics are provided in the appendix. These analyses further examine how the explicit associative-memory mechanisms contribute to the robustness, efficiency, and behavior of V-HMN.

\begin{table*}[t]
\centering
\caption{Comparison of V-HMN with baseline models on CIFAR-10, CIFAR-100, 
SVHN, and Fashion-MNIST. Top-1 test accuracy (\%) are reported as mean $\pm$ standard deviation over 3 seeds.
Baselines include ViT \citep{dosovitskiy2020vit}, 
Swin-ViT \citep{liu2021swin}, 
MLP-Mixer \citep{tolstikhin2021mlpmixer}, 
MetaFormer \citep{yu2022metaformer}, 
Vim \citep{zhu2024visionmamba}, 
and AiT \citep{sun2025associative}.} 
\label{tab:aug-comparison}
\resizebox{0.85\textwidth}{!}{
\begin{tabular}{l|c|c|c|c|c}
\toprule
\textbf{Model} & \textbf{CIFAR-10} & \textbf{CIFAR-100} & \textbf{SVHN} & \textbf{FashionMNIST} & \textbf{Params (M)} \\
\midrule
ViT                 & $91.66_{\pm 0.08}$ & $72.56_{\pm 0.01}$   & $96.11_{\pm 0.29}$  & $91.83_{\pm 0.15}$  & 7.16 \\
Swin-ViT            & $87.94_{\pm 0.10}$  & $66.58_{\pm 0.37}$   & $95.89_{\pm 0.17}$  & $91.79_{\pm 0.17}$  & 6.92  \\
MLP-Mixer           & $92.65_{\pm 0.26}$  & $73.35_{\pm 0.39}$   & $96.91_{\pm 0.06}$ & $91.46_{\pm 0.32}$  & 8.71 \\
MetaFormer          & $75.40_{\pm 1.56}$  & $48.76_{\pm 0.75}$   & $86.54_{\pm 2.71}$  & $91.42_{\pm 0.21}$  & 6.99  \\
Vim        & $82.83_{\pm 3.87}$  & $63.29_{\pm 2.83}$   &  $86.63_{\pm 1.51}$ &  $88.63_{\pm 1.49}$ & 7.12  \\ 
AiT  & $92.97_{\pm 0.30}$  & $72.91_{\pm 0.17}$  & $95.98_{\pm 0.06}$  & $91.51_{\pm 0.04}$  & 7.15 \\
V-HMN (ours)        & {$\boldsymbol{93.94_{\pm 0.11}}$} & {$\boldsymbol{76.58_{\pm 0.09}}$} & {$\boldsymbol{97.16_{\pm 0.04}}$} & {$\boldsymbol{92.27_{\pm 0.06}}$} & 7.12 \\
\bottomrule
\end{tabular}} 
\end{table*}

\begin{table}[t]
\centering
\caption{Comparison on ImageNet-1k. All results are Top-1 accuracy (\%). ($\dagger$) indicates our re-implementation under the same setting for fair comparison.}
\vspace{1ex}
\label{tab:imagenet_comparison}
\resizebox{0.57\linewidth}{!}{
\begin{tabular}{lccc}
\toprule
\textbf{Method} & \textbf{Image Size} & \textbf{\#Params} & \textbf{Top-1 Acc.} \\
\midrule
ResNet-50 \citep{he2016resnet} & 224 & 26M & 76.1 \\ 
SENet-50 \citep{hu2018squeeze} & 224 & 28M & 76.7 \\
\midrule
ViT-B/16 \citep{dosovitskiy2020vit} & 384 & 86M & 77.9 \\  
ViT-B/16$^{\dagger}$ \citep{dosovitskiy2020vit} & 224 & 86M & 76.2 \\  
ViT-L/16 \citep{dosovitskiy2020vit} & 384 & 307M & 76.5 \\ 
MLP-Mixer-B/16 \citep{tolstikhin2021mlpmixer} & 224 & 59M & 76.4 \\   
\midrule
V-HMN (ours) & 224 & 26M & 76.7 \\  
\bottomrule
\end{tabular}
}   
\end{table}

\subsection{Main Results}

Table~\ref{tab:aug-comparison} reports the comparison of V-HMN with a wide range of mainstream vision backbones, including Transformer-based models (ViT, Swin-ViT), MLP-based architectures (MLP-Mixer, MetaFormer), state-space models (Vim), and the recently proposed AiT. 
Despite having a comparable parameter scale, V-HMN consistently achieves the best performance across all benchmarks, reaching $93.94\%$ on CIFAR-10, $76.58\%$ on CIFAR-100, $97.16\%$ on SVHN, and $92.27\%$ on Fashion-MNIST. 

We attribute these improvements to two key design choices. 
First, the incorporation of \emph{local and global associative memories} enables the model to retrieve and integrate prototypical patterns, effectively supplementing limited supervision with reusable priors. 
Second, the \emph{iterative refinement mechanism} provides a lightweight error-corrective process that gradually aligns representations with memory-predicted prototypes, thereby enhancing the robustness. 
Compared with standard feedforward Transformers or purely sequential state-space models, V-HMN benefits from persistent, content-addressable prototypes that capture both local details and global context, yielding stronger generalization under comparable model capacity. 

A key observation is that V-HMN surpasses AiT \citep{sun2025associative} across all datasets, despite having nearly identical parameter counts. 
While AiT integrates associative memory \emph{within} a Transformer layer, its token mixing remains attention-centric. 
In contrast, V-HMN is \emph{memory-centric}: local and global Hopfield modules replace self-attention as the mixing mechanism, memories are persistent and class-balanced (written with real sample embeddings during training and frozen at inference), and representations are updated through a predictive-coding-inspired refinement loop. 
These results suggest that making associative memory the \emph{primary} computational primitive, rather than an auxiliary component to Transformer, provides better data efficiency and accuracy under comparable model capacity.

To evaluate whether the proposed memory-centric design generalizes beyond small-scale benchmarks, we conduct experiments on ImageNet-1k and report the results in Table~\ref{tab:imagenet_comparison}. 
Rather than targeting state-of-the-art performance on this large-scale benchmark, our goal is to assess whether the core associative memory mechanism remains effective when scaled to higher-resolution inputs and more complex data distributions. We observe that a relatively small V-HMN model achieves 76.7\% Top-1 accuracy with only 26M parameters, remaining competitive with substantially larger Transformer-based backbones under comparable input resolutions. 
This suggests that the proposed memory-based retrieval and refinement mechanism remains viable at ImageNet scale and can serve as a parameter-efficient alternative to standard token-mixing strategies in this setting. 
Importantly, V-HMN does not incorporate architectural components commonly used in modern large-scale vision backbones, such as multi-scale designs~\citep{wang2021pvt}. This design choice is intentional, as it allows us to isolate the effect of the memory-based computation without relying on additional engineering optimizations. Despite this simplified architecture and minimal tuning, the model remains effective at ImageNet scale, indicating that the underlying inductive bias is robust. 
Beyond performance, V-HMN maintains a transparent computational structure, where predictions are explicitly influenced by retrieved memory patterns and iterative refinement. This property is difficult to preserve in more complex architectures and provides a foundation for analyzing how stored representations contribute to the final decision.

Appendix~\ref{appendix:memory-bank-design-controls} extends the main accuracy results by comparing class-balanced ring buffers, unbalanced ring buffers, learned prototypes, and EMA prototypes under the same 400-epoch CIFAR-10/CIFAR-100 training settings, showing that V-HMN is not tied to a single memory construction. Appendix~\ref{appendix:complexity-runtime-analysis} reports the corresponding compute, throughput, GPU-memory, and stored-memory measurements, separating trainable weights from stored latent memory entries and identifying neighborhood gathering and Hopfield-style retrieval as measurable runtime components.

\begin{figure*}[t]
    \centering
    \includegraphics[width=\linewidth]{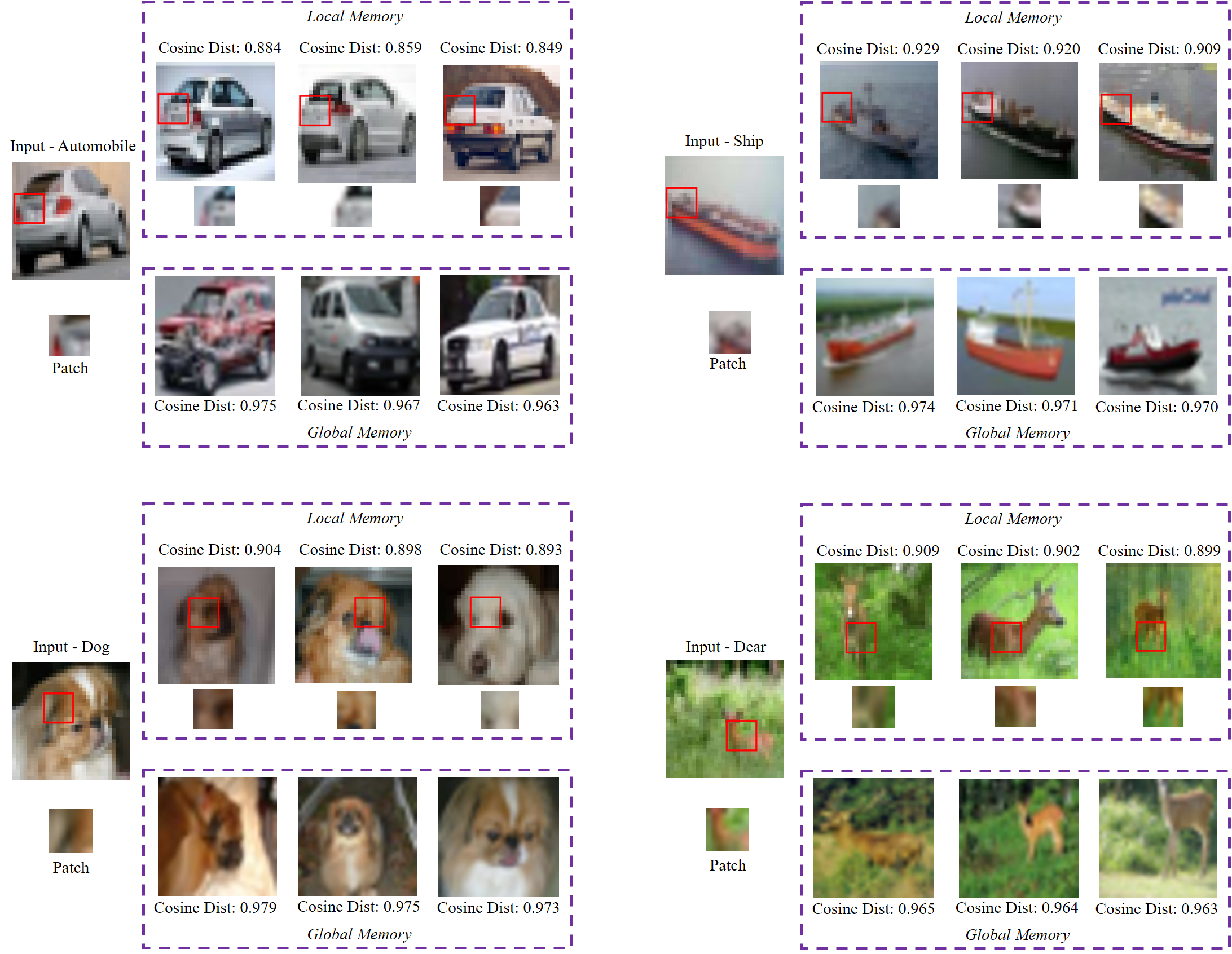}
    \caption{Visualization of retrieved prototypes from local and global memory.} 
    \label{fig:vis} 
    \vspace{-1ex}
\end{figure*}

\subsection{Visualization}

To better understand the behavior of V-HMN, we visualize the retrieved prototypes from both local and global memory banks. 
For each test image, we first identify the most informative patch based on the attention pooling weights (highlighted by the red box), and then retrieve its nearest prototypes from the corresponding memory banks. 
To clearly illustrate the retrieved regions, we adopt a patch size of $8\times 8$ during visualization.

Figure~\ref{fig:vis} shows representative results across four categories: automobile, ship, dog, and deer. 
Several interesting observations can be made. 
First, the \emph{local memory} retrieves highly similar local structures from other images of the same class.  
For example, in the automobile and ship cases, the retrieved prototypes consistently align to similar positions, highlighting that local memory captures part-level correspondences across samples. 
This demonstrates that local memory is capable of associating fine-grained visual parts across different samples, thereby enhancing spatial interpretability. 
Second, the \emph{global memory} retrieves holistic prototypes that provide complementary, scene-level priors. For instance, as shown in the dog and deer cases, the global retrieval captures diverse poses and viewpoints of the same class, which supply additional information to complete or refine the local representation. 
This behavior is consistent with the role of associative memory, which not only recalls similar exemplars but also provides missing context to stabilize predictions.

Together, these results highlight that the two memory paths confer complementary benefits: local memory aligns semantically corresponding parts across samples, while global memory furnishes broader priors that help disambiguate incomplete or noisy inputs. 
Such explicit retrieval and refinement provide a clear and inspectable view of the prototype-based reasoning process in V-HMN, as one can directly inspect which prototypes are retrieved for a given input. To test whether these retrieval traces are functionally relevant rather than only visual examples, Appendix~\ref{appendix:retrieval-disruption-tests} evaluates paired inference-time interventions that disable or perturb populated local and global memory reads. These interventions change predictions, supporting retrieval-based prototype transparency as a behavioral property of the model.


\section{Summary and Outlook}In this work, we introduced \textbf{V-HMN}, a brain-inspired vision backbone that organizes each block around local and global Hopfield-style memory. Through associative retrieval and predictive-coding-inspired refinement, V-HMN moves beyond purely feedforward or self-attention architectures and places memory at the center of feature integration. This yields two key benefits: \emph{data efficiency}, by reusing stored prototypes as inductive priors, and \emph{interpretability}, 
as retrieved prototypes provide an inspectable trace of the patterns involved in each prediction.  V-HMN outperforms vision backbones of comparable scale across CIFAR, SVHN, and Fashion-MNIST, and further demonstrates that its inductive bias remains effective when scaled to ImageNet, achieving reasonable performance without large-scale architectural tuning or extensive hyperparameter optimization.

Our study highlights the promise of memory-centric backbones as a viable and scalable alternative to mainstream vision backbones. By grounding representation learning in explicit associative memory and lightweight error-corrective refinement, V-HMN bridges biologically inspired computation with large-scale machine learning. While this work focuses on image classification, the underlying memory-driven principle may extend to other vision tasks, such as retrieval and metric learning, few-shot adaptation, and dense prediction problems including segmentation and detection, where such a principle could provide beneficial inductive biases. 
We believe these directions open the door to more interpretable, data-efficient, and biologically inspired architectures.
 
\begin{ack}
This research was funded in part by the Austrian Science Fund (FWF) 10.55776/COE12 and the AXA Research Fund. Amine M'Charrak gratefully acknowledges support from the Evangelisches Studienwerk e.V.\ Villigst through a doctoral fellowship. 
\end{ack}

\bibliographystyle{plain}
\bibliography{references}


\clearpage
\appendix

\section{Appendix}

\subsection{Hierarchical Local--Global Memory and Invariances}
\label{appendix:hierarchy-invariances}

V-HMN is organized as a stack of HMN blocks, each equipped with its own local and global memory banks.
The hierarchical structure arises because each block processes and stores representations at its own depth: the prototypes learned at lower layers are grounded in early, fine-grained features, while deeper layers operate on progressively transformed and more semantically organized representations produced by the preceding blocks.
As a result, lower layers learn prototypes of local edge- and texture-like patterns, whereas higher layers learn more abstract object- and scene-level prototypes.

\paragraph{Layerwise interaction of local and global memories.}
At every layer, the local HMN operates on overlapping neighborhoods of the token grid to retrieve and refine local patterns, while the global HMN pools information across all tokens to retrieve a scene-level prototype that is broadcast back to the entire layer.
Importantly, each layer’s memory banks only interact with the representations produced at that same depth; deeper layers never directly access shallow-layer features.
This architectural separation induces a genuine hierarchy of prototypes.
Lower layers store and reinstate fine-grained edge-, texture-, and patch-level patterns, whereas higher layers operate on increasingly abstract representations passed up from previous blocks.
The retrieved global prototype at each layer provides a contextual prior that guides local refinement, enabling ambiguous or partially occluded patches to be interpreted in a way consistent with the overall scene.
Through this layerwise progression, V-HMN builds increasingly holistic and context-aware representations without relying on explicit spatial downsampling.

\paragraph{What invariances the hierarchy provides.}
The hierarchical local--global design induces several useful invariances, but not all invariances observed in practice come from the architecture alone.

\begin{itemize}
  \item \textbf{Local tolerance to small perturbations.} Because local neighborhoods are unfolded with overlap, a small translation or deformation of a feature (e.g., shifting an edge by one pixel) changes which tokens contribute to a neighborhood but often leaves its nearest local prototype unchanged. The local Hopfield retrieval therefore tends to map slightly perturbed patches back to the same or a nearby prototype, providing robustness to small local jitters and noise.

  \item \textbf{Contextual invariance via global prototypes.} The global HMN sees a pooled summary of the entire token grid. As a result, global prototypes are largely insensitive to where an object appears within the image, acting more like a translation-tolerant scene or object code. When the global prototype is broadcast back to all tokens, it stabilizes local representations against clutter or partial occlusion: different arrangements of the same global configuration are attracted to similar global memory slots.

  \item \textbf{Increasing invariance with depth.} As representations become progressively more abstract across successive HMN blocks, prototypes in deeper local and global memory banks become less sensitive to fine-grained pixel-level details and more sensitive to object- and class-level structure. This yields a degree of scale and translation tolerance at higher layers, analogous to the progression observed along the ventral visual stream.
\end{itemize}

\paragraph{What is handled by augmentation, preprocessing, and pooling.}
Several important invariances are primarily provided by standard deep-learning components rather than by the memory hierarchy itself.
Random crops, horizontal flips, and optional AutoAugment are the main sources of robustness to large translations, flips, and complex photometric distortions.
The patch embedding and final attention pooling contribute additional translation invariance by making the classifier depend mostly on aggregated token statistics rather than exact pixel coordinates.
We do not build in explicit rotation or scale-equivariant structure; robustness to such transformations arises empirically from data augmentation and from the generic effects of depth and pooling, not from a specialised architectural mechanism.

\medskip
In summary, the hierarchical arrangement of local and global memory banks provides a structured inductive bias: lower layers learn local prototypes that are robust to small perturbations; higher layers and global memories learn more holistic prototypes that are tolerant to object location and clutter. This interacts synergistically with standard augmentation and pooling to produce the overall invariance profile observed in our experiments.

\subsection{Variance of cosine similarity}
\label{appendix:variance-proof}

Let $\hat q,\hat k \in \mathbb{R}^D$ be independent random unit vectors. We assume the random unit vectors are drawn from a distribution that is invariant under coordinate permutations and sign flips (e.g., the uniform distribution on the unit sphere). 
We first compute the second moment of $\hat q$. 
Write $\hat q=(\hat q_1,\dots,\hat q_D)$. By coordinate symmetry, all coordinates satisfy 
$\mathbb{E}[\hat q_i^2]=v$. Since $\sum_{i=1}^D \hat q_i^2=1$, taking expectations gives
\[
Dv=1 \;\;\Rightarrow\;\; v=\tfrac{1}{D}.
\]
For $i\neq j$, flipping the sign of the $i$-th coordinate (which preserves the distribution of a random unit vector)
implies
\[
\mathbb{E}[\hat q_i \hat q_j] = -\,\mathbb{E}[\hat q_i \hat q_j] \;\;\Rightarrow\;\; \mathbb{E}[\hat q_i \hat q_j]=0.
\]
Hence the off-diagonal entries vanish and the diagonal entries are all $1/D$, i.e., 
\[
\mathbb{E}[\hat q\,\hat q^\top] = \tfrac{1}{D} I.
\]

It follows that
\[
\mathbb{E}[\hat q^\top \hat k]
=\mathbb{E}_{\hat q}\!\big[\hat q^\top\,\mathbb{E}_{\hat k}[\hat k]\big]=0.
\]
For the second moment,
\begin{align*}
\mathbb{E}\!\big[(\hat q^\top \hat k)^2\big]
&=\mathbb{E}_{\hat k}\!\Big[\hat k^\top\,\mathbb{E}_{\hat q}[\hat q\,\hat q^\top]\,\hat k\Big] \\
&=\mathbb{E}_{\hat k}\!\Big[\hat k^\top \Big(\tfrac{1}{D}I\Big)\hat k\Big] \\
&=\tfrac{1}{D}\,\mathbb{E}_{\hat k}[\|\hat k\|_2^2] \\
&=\tfrac{1}{D}.
\end{align*}
Therefore,
\[
\operatorname{Var}(\hat q^\top \hat k)
=\mathbb{E}\!\big[(\hat q^\top \hat k)^2\big]
-\big(\mathbb{E}[\hat q^\top \hat k]\big)^2
=\tfrac{1}{D}.
\]

Concluding, the variance of the cosine similarity between two random unit vectors
decays as $1/D$. Multiplying the similarity by $\sqrt{D}$ rescales the logits to 
approximately unit variance before the softmax.
 
\subsection{Iterative refinement as predictive-coding (PC) dynamics} 
\label{appendix:refinement-pc}

The central operation in both local and global modules is an iterative refinement rule. Given a
current representation $z^{(t)} \in \mathbb{R}^D$ and a memory bank $M \in \mathbb{R}^{K \times D}$ with rows $M_j \in \mathbb{R}^D$,
we first perform Hopfield-style retrieval as described in Section~3.2:
\begin{align}
  \hat z^{(t)} &= \frac{z^{(t)}}{\lVert z^{(t)} \rVert_2}, \quad
  \hat M_j = \frac{M_j}{\lVert M_j \rVert_2}, \nonumber\\
  \alpha_j^{(t)} &= \mathrm{softmax}_j\!\big(\sqrt{D}\,\hat z^{(t)} \hat M_j^\top\big), \nonumber\\
  m^{(t)} &= \sum_{j=1}^K \alpha_j^{(t)} M_j ,
  \label{eq:retrieve}
\end{align}
where $M_j$ denotes the $j$-th memory slot, and $m^{(t)}$ is the retrieved memory. We then update
$z^{(t)}$ via
\begin{equation}
  z^{(t+1)} \;=\; z^{(t)} + \beta \big(m^{(t)} - z^{(t)}\big),
  \label{eq:refinement}
\end{equation}
where $\beta$ is a learnable update strength.

This rule can be interpreted as a simple predictive-coding (PC) dynamics. Define the local
\emph{prediction error}
\begin{equation*}
  \varepsilon^{(t)} \;:=\; m^{(t)} - z^{(t)} .
\end{equation*}
Then, Eq.~\ref{eq:refinement} becomes
\begin{equation*}
  z^{(t+1)} \;=\; z^{(t)} + \beta\,\varepsilon^{(t)},
\end{equation*}
which is a discrete gradient-descent step on the squared prediction-error energy
\begin{equation*}
  \mathcal{F}(z) \;=\; \tfrac{1}{2}\,\lVert z - m(z) \rVert_2^2 .
\end{equation*}
If we treat $m(z)$ as fixed with respect to $z$ during one update step, then
\begin{equation*}
  \nabla_z \mathcal{F}(z) \;\approx\; z - m(z) \;=\; -\,\varepsilon^{(t)},
\end{equation*}
and Eq.~\ref{eq:refinement} coincides with
\begin{equation*}
  z^{(t+1)} \;\approx\; z^{(t)} - \beta \nabla_z \mathcal{F}(z^{(t)}).
\end{equation*}
That is, the Hopfield module produces a memory-based prediction $m^{(t)}$ of the code
$z^{(t)}$, the residual~$\varepsilon^{(t)}$ acts as a prediction-error signal, and the representation is
iteratively corrected, so as to minimize a local prediction-error energy.

This mirrors the core mechanism of hierarchical predictive coding \citep{rao1999predictive,friston2005theory,
whittington2019theories}: higher-level causes generate a prediction of lower-level activity, explicit
error units encode their difference, and latent states are updated by gradient descent on a
prediction-error or free-energy objective. In V-HMN, the role of the generative model is played by
the associative memory: memory slots $M_j$ act as prototypical causes, the similarity scores
$\alpha_j^{(t)}$ approximate a posterior over these causes given $z^{(t)}$, and the retrieved prototype
$m^{(t)} = \mathbb{E}[M_j \mid z^{(t)}]$ provides the top-down prediction. The refinement dynamics
of Eq.~\ref{eq:refinement} thus implements a lightweight PC update in which (i) the
\emph{feedforward} path computes similarities and posterior weights from the current features to the
memory slots, and (ii) the \emph{feedback} path injects the memory-based prediction back into the
features through the prediction-error signal $(m^{(t)} - z^{(t)})$. In practice, we find that one or two
iterations are sufficient to obtain robust improvements while keeping computation efficient.

\subsection{Biological plausibility and relation to cortical circuits}
\label{appendix:bio-plausibility}

V-HMN is not intended as a faithful anatomical model of the primate visual system, but it is
explicitly guided by two computational motifs that are widely discussed in systems neuroscience:
(i) associative  memory implemented by recurrently connected populations, and
(ii) predictive-coding-like error-corrective refinement in cortical microcircuits. We now 
clarify where our design aligns with known hippocampal and cortical circuitry and where the
connection is only analogical.

\paragraph{Associative memory beyond the hippocampus.}
Associative dynamics are often introduced via models of the hippocampal formation,
where pattern separation in dentate gyrus and pattern completion in CA3 are thought to
support episodic memory and spatial navigation. However, a large body of theoretical and
experimental work suggests that related forms of autoassociative computation are also
implemented in neocortical microcircuits, where recurrent collateral connections between
pyramidal neurons can sustain stable activity patterns that function as long-term and short-term
memories, perceptual representations, and decision states. In this broader view, associative
memory is a \emph{general} cortical computational motif, not something confined to the
hippocampus. Our local Hopfield modules are designed to echo this idea: they implement
content-addressable retrieval over local patch-level embeddings using modern Hopfield
dynamics, such that frequently co-occurring visual features (e.g., edges, corners, textures)
form structured prototype-like representations stored in the memory bank. These modules
enable V-HMN to integrate local contextual information in a manner that is both
interpretable and robust, supporting refinement steps that adjust features toward semantically consistent local patterns, even under perturbations.

\paragraph{Global memory and hippocampal/entorhinal inspiration.}
At the same time, the global Hopfield module and its class-balanced episodic memory bank are
more directly inspired by hippocampal and entorhinal circuitry. The global query aggregates
information across the entire image and retrieves a scene-level prototype from a separate memory
bank, loosely analogous to how hippocampus and entorhinal cortex integrate inputs from many
cortical areas into a sparse episodic code that can later be reinstated to bias cortical activity.
The broadcast of the retrieved global prototype back to all tokens in a block is therefore
reminiscent of hippocampo--cortical feedback that reinstates context or episodes, but at a highly
abstract level and without any claim of anatomical fidelity.

\paragraph{Relation to ventral visual stream processing.}
In the brain, visual object recognition emerges along the ventral visual stream (V1, V2, V4, IT), with rich recurrent connectivity and local microcircuits, before hippocampal structures become involved in binding objects into episodic memories.
V-HMN compresses some of these ideas into a single backbone: local HMN blocks can be viewed as abstracted cortical microcircuits that combine feedforward feature extraction with local associative retrieval and predictive-coding-style refinement, and the global HMN adds an additional, hippocampus-inspired contextual signal.
We do not claim a one-to-one mapping between layers of V-HMN and specific cortical areas.
Rather, our aim is to capture, within a practical vision model, the core computational principles of (i) distributed pattern reinstatement through associative memory and (ii) iterative interaction between top-down predictions and bottom-up features.

\paragraph{Brain-inspired, not anatomically faithful.}
Taken together, these design choices place V-HMN somewhere between conventional deep
vision backbones and detailed biophysical models. Compared to standard feedforward or
self-attention-only architectures, V-HMN moves closer to known cortical and hippocampal
computation by making associative memory and error-corrective dynamics first-class citizens in
the backbone: memory banks correspond to explicit, inspectable prototypes rather than hidden
weights; Hopfield retrieval implements pattern completion; and iterative refinement
reduces local prediction errors. At the same time, we abstract away from many anatomical
details (layer-specific connectivity, cell types, precise ventral-stream staging), and we do not
claim that V-HMN is a literal model of the primate visual system. Instead, our goal is to move
towards more biologically grounded computation than in current deep learning models, while
remaining competitive and scalable on standard machine-learning benchmarks.

\subsection{Why V-HMN is data-efficient}
\label{appendix:data-efficiency-analysis}

The empirical results in Tables~\ref{tab:data-fractions} and~\ref{tab:data-efficiency-all-models} show that
V-HMN maintains strong performance even when only a small fraction of the training labels is
available. For instance, with just 10\% of the labeled data, V-HMN achieves $80.22\%$ on
CIFAR-10, $43.21\%$ on CIFAR-100, and $89.18\%$ on Fashion-MNIST
(Table~\ref{tab:data-fractions}), and it consistently outperforms ViT, Swin-ViT, MLP-Mixer,
MetaFormer, Vim, and AiT under both 10\% and 30\% training
data across all three benchmarks (Table~\ref{tab:data-efficiency-all-models}). We now briefly analyze why
V-HMN is more data-efficient than these baselines.

\paragraph{Prototype-based nonparametric prior.}
Both the local and global Hopfield modules retrieve representations from explicit memory banks whose slots are populated with real latent embeddings during training. These slots act as prototypes that approximate class-conditional manifolds in feature space. Once a prototype is stored, future inputs can leverage it via retrieval even if supervision is limited, providing a nonparametric prior that complements the parametric backbone. In low-data regimes, this prototype reuse compensates for limited gradient-based fitting and reduces overfitting to the small labeled set. The monotonic improvements from 10\% to 30\% to 50\% labeled data in Table~\ref{tab:data-fractions} reflect this behavior: as more labeled examples are observed, memory banks become better populated and the same prototypes can be reused across many inputs.

\paragraph{Local--global inductive bias.}
V-HMN enforces a structured inductive bias by combining (i) local Hopfield dynamics on unfolded
$k \times k$ neighborhoods and (ii) a global Hopfield module operating on scene-level aggregates.
Local memories specialize to recurring edge- and texture-like patches, while global memories
capture higher-level scene and object prototypes. This hierarchical organization constrains the
effective hypothesis class: new images are explained in terms of reusing and recombining a finite
set of learned local and global prototypes, rather than learning fresh features from scratch for every
configuration. This reduces the amount of labeled data needed to reach a given accuracy, which is
consistent with the stronger gains of V-HMN in the 10\% and 30\% settings compared to the
full-data regime in Table~\ref{tab:data-fractions}.

 \paragraph{Iterative refinement focuses capacity on hard examples.}
As discussed in Section~\ref{appendix:refinement-pc}, the refinement rule implements a predictive-coding (PC) update, where representations are iteratively corrected to reduce local prediction-error energy. 
In practice, this means that model capacity is concentrated on those regions of feature space where the memory-based predictions and current representations disagree. 
When data are scarce, this mechanism helps stabilize learning: easy examples quickly align with their nearest prototypes and require little further adjustment, 
while scarce or atypical examples receive larger refinement updates (proportional to prediction error). 
The ablation in Table~\ref{tab:iteration_ablation} shows that removing refinement ($t=0$) substantially hurts performance and that one or two refinement iterations yield the best trade-off between accuracy and computation. 
Together with the memory-based priors above, this targeted refinement explains why V-HMN achieves higher accuracies than standard backbones and AiT under limited supervision (Table~\ref{tab:data-efficiency-all-models}), despite using a comparable number of parameters.

\subsection{Implementation Details} 
V-HMN is implemented in PyTorch. As for the experiments on small datasets (CIFAR, Fashion-MNIST, and SVHN), images are divided into non-overlapping $4\times4$ patches and embedded into tokens with a learnable positional encoding. The backbone consists of six stacked V-HMN blocks, each equipped with a local Hopfield module operating on $3\times3$ spatial neighborhoods and a global Hopfield module operating on the mean-pooled query. The local and global modules maintain class-balanced memory banks with 2500 and 1000 slots, respectively.  The embedding and latent dimensions are both set to 256, and the MLP expansion ratio is set to 2. Refinement updates are controlled by a learnable parameter $\beta$ initialized to 0.2, and unless otherwise specified, a single refinement iteration is used.
We use the Adam optimizer \citep{kingma2014adam} with cosine learning-rate decay. The initial learning rate is 0.001, linearly warmed up during the first five epochs. Training is conducted for 400 epochs with a batch size of 256 and weight decay of $5\times10^{-5}$. For data augmentation, we adopt a strong augmentation including random cropping, horizontal flipping, AutoAugment \citep{cubuk2019autoaugment}, MixUp \citep{zhang2017mixup}, and CutMix \citep{yun2019cutmix}. All reported results are based on this training setup unless otherwise noted.

As for the ImageNet experiments, we use a batch size of 1000 and a patch size of $14\times14$. 
The local and global memory banks contain 5000 and 3000 slots, respectively.
The model adopts a four-stage hierarchical design with embedding and latent dimensions of $(128, 224, 320, 512)$ across stages and depths of $(3, 3, 4, 2)$, while the MLP expansion ratio is set to 4.
We train the model for 310 epochs using the AdamW optimizer with a cosine learning-rate schedule.
The initial learning rate is 0.001 with linear warmup at the beginning of training.
A weight decay of 0.03 is applied, and gradient clipping with a maximum norm of 0.1 is used.
For data augmentation, we use RandAugment, MixUp, CutMix, and random erasing.
We do not use stochastic depth, repeated augmentation, or Exponential Moving Average (EMA).
Training is performed using 4 NVIDIA H100 GPUs.

\subsection{Retrieval-Disruption Tests}
\label{appendix:retrieval-disruption-tests}

This appendix reports 400-epoch controlled analyses of the V-HMN memory system on CIFAR-10 and CIFAR-100. Compact table labels are defined here: CB-ring is the class-balanced ring buffer, U-ring is the unbalanced ring buffer, EMA is exponential moving average, proto. means prototype, mem. means memory, FLOPs means floating-point operations, GPU mem. means peak allocated GPU memory, and Top-1/Top-5 are classification accuracy metrics.

The retrieval-disruption analysis separates the 400-epoch training recipe from the sample-memory state used at evaluation. In V-HMN, local and global retrieval read from memory banks of latent features rather than from raw images. The local bank stores projected local-neighborhood features selected from training images, while the global bank stores projected image-level features obtained from mean-pooled token representations. During ordinary supervised training, these detached feature vectors are written into memory slots associated with their class labels; at evaluation, the memory banks are frozen and the model reads from them without using labels. Retrieval interventions are therefore meaningful only when those sample memories contain class-labeled feature entries. Clean labels denotes training without MixUp/CutMix, so ordinary class labels are available for memory writes throughout training. Strong aug. denotes the main CIFAR recipe from the implementation details: random crop, horizontal flip, AutoAugment, MixUp, and CutMix. In that mixed-label path, MixUp/CutMix targets are valid for the classification loss but are not written into class-indexed sample memories. Thus, the Strong aug. rows in Table~\ref{tab:full400_retrieval_disruption} report the accuracy of the main strong-augmentation recipe, but their sample memories are not populated and local/global memory-intervention cells are not applicable. Strong aug.+clean mem. uses the same strong-augmentation objective and schedule and adds a clean-label no-gradient pass after each epoch to populate the local and global sample memories with ordinary class-labeled feature entries. For rows with populated sample memory, we evaluate paired inference-time interventions in identical sample order with frozen memory banks and no labels during evaluation. The modes include standard retrieval, disabling local/global/both retrieval, feature-shuffling memory vectors, and random top-$k$ retrieval. These interventions quantify how retrieval changes predictions; because attention pooling aggregates token information before the final classifier, retrieved prototypes are best interpreted as inspectable retrieval traces rather than complete class-level rationales.

\input{generated_results/tab_full400_retrieval_disruption}

Table~\ref{tab:full400_retrieval_disruption} is the sole quantitative summary for the retrieval-disruption analysis. Each intervention cell has the form Top-1 drop / changed predictions, where Top-1 drop is measured in percentage points against that same row's Standard Top-1, and changed predictions is the percentage of test examples whose predicted class changes under the intervention. The table separates three behaviors. Clean-label models have populated sample memories and show the largest paired intervention effects. Strong aug. matches the main strong-augmentation training policy and achieves high accuracy, but its sample memories are not populated because mixed-label batches are not written into class-indexed memory banks. These entries are therefore marked as not applicable rather than reported as numerical intervention measurements. Strong aug.+clean mem. keeps the same 400-epoch strong-augmentation objective and schedule, adds a clean no-gradient memory-population pass, and restores measurable paired retrieval effects, especially for local retrieval and memory perturbations. The Strong aug. and Strong aug.+clean mem. rows should be read as a controlled memory-state comparison under the same training objective, not as evidence that adding a memory-population pass is a new augmentation method.

\FloatBarrier

\subsection{Analysis of Different Memory Constructions}
\label{appendix:memory-bank-design-controls}

\begin{figure}[!htbp]
    \centering
    \includegraphics[width=\linewidth]{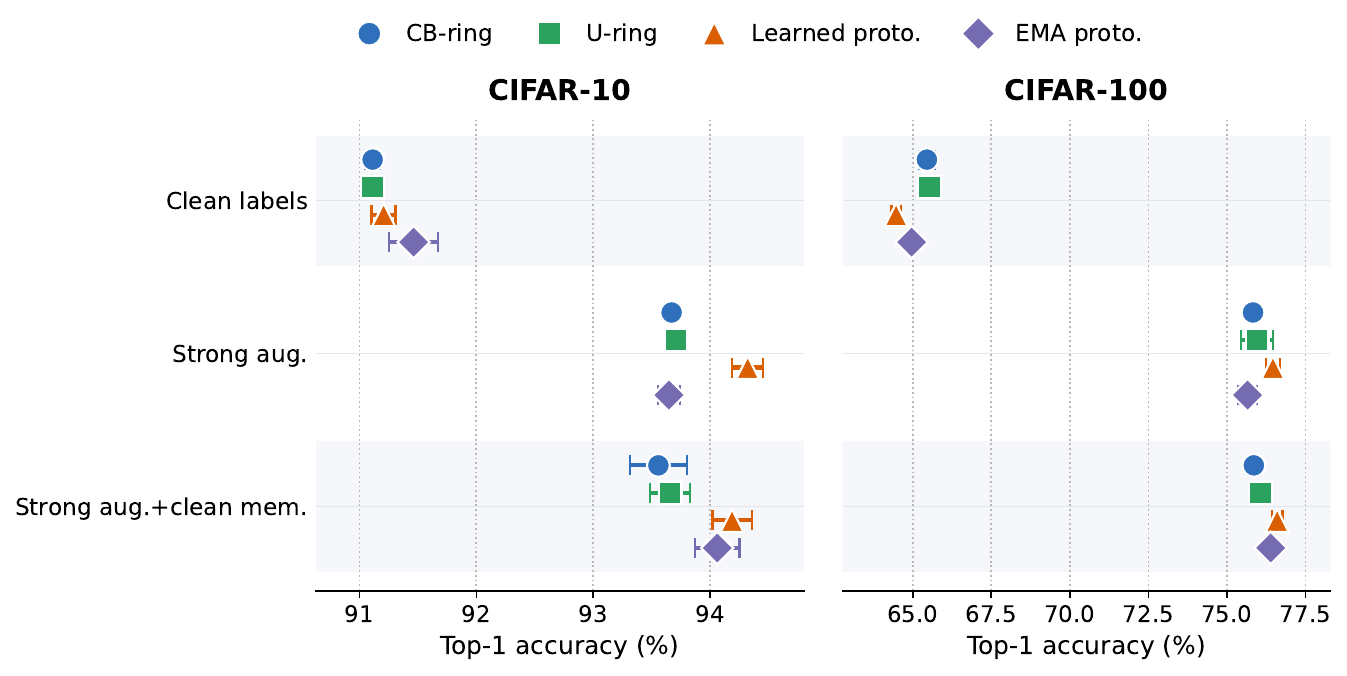}
    \caption{Top-1 accuracy for matched memory constructions under 400-epoch training. Points show the mean over three independent runs and horizontal bars show one standard deviation. Each panel compares the same four memory constructions under the same model, slot budget, latent dimension, retrieval rule, and optimizer. The figure focuses on Top-1 because Top-5 is near-saturated for these CIFAR settings; storage mechanics are summarized in Table~\ref{tab:full400_memory_bank_properties}.}
    \label{fig:full400_memory_bank_top1}
\end{figure}

The memory-bank analysis compares class-balanced, unbalanced, learned-prototype, and EMA-prototype memories under the same architecture, local/global slot budget, latent dimension, retrieval rule, and optimizer within each dataset. Unless otherwise stated, the main-paper V-HMN experiments use the class-balanced ring buffer: detached projected features from labeled training examples are written into class-indexed local and global memory slots, with a fixed per-class capacity. The global entries correspond to image-level latent features from training examples, but they are not raw training images. The unbalanced ring buffer keeps the same example-written storage capacity but removes per-class balancing. Learned prototypes keep the same retrieval interface and local/global slot counts, but the memory entries are free trainable latent vectors optimized by back-propagation rather than features copied from particular training examples. EMA prototypes are not learned by back-propagation: for each class, the local and global prototype is initialized from observed class features and then updated as an exponential moving average of later class features. Thus, learned-prototype and EMA-prototype banks test alternative memory semantics under the same Hopfield-style read operation, while CB-ring and U-ring test different ways of storing example-derived features. Figure~\ref{fig:full400_memory_bank_top1} reports the 400-epoch CIFAR-10 and CIFAR-100 Top-1 accuracy results for three memory conditions: clean-label training with ordinary supervised memory writes, strong augmentation where mixed-label batches train the classifier but do not populate class-indexed sample memory, and strong augmentation with the clean no-gradient memory-population pass used for populated-memory retrieval analyses. Table~\ref{tab:full400_memory_bank_properties} summarizes the corresponding memory construction and storage properties.

\FloatBarrier

\input{generated_results/tab_full400_memory_bank_properties}

The split presentation separates accuracy from storage mechanics. Clean-label training isolates ordinary supervised memory population, Strong aug. measures the main mixed-label augmentation recipe without class-indexed sample-memory writes, and Strong aug.+clean mem. restores sample-memory reads for the retrieval analyses while preserving the same training objective and schedule. Across both datasets, learned-prototype and EMA-prototype memories remain competitive with ring-buffer memories, supporting the broader explicit-memory inductive bias rather than a single storage construction. The storage table also explains why EMA uses fewer entries: it stores one local and one global class-mean feature prototype per class, whereas the ring-buffer and learned-prototype designs keep the matched 2500 local and 1000 global memory entries. Consequently, EMA prototypes summarize classes in latent feature space; they are not individual examples and do not correspond to specific training images.

\FloatBarrier

\subsection{Runtime and Memory Analysis}
\label{appendix:complexity-runtime-analysis}

\begin{figure}[!htbp]
    \centering
    \includegraphics[width=\linewidth]{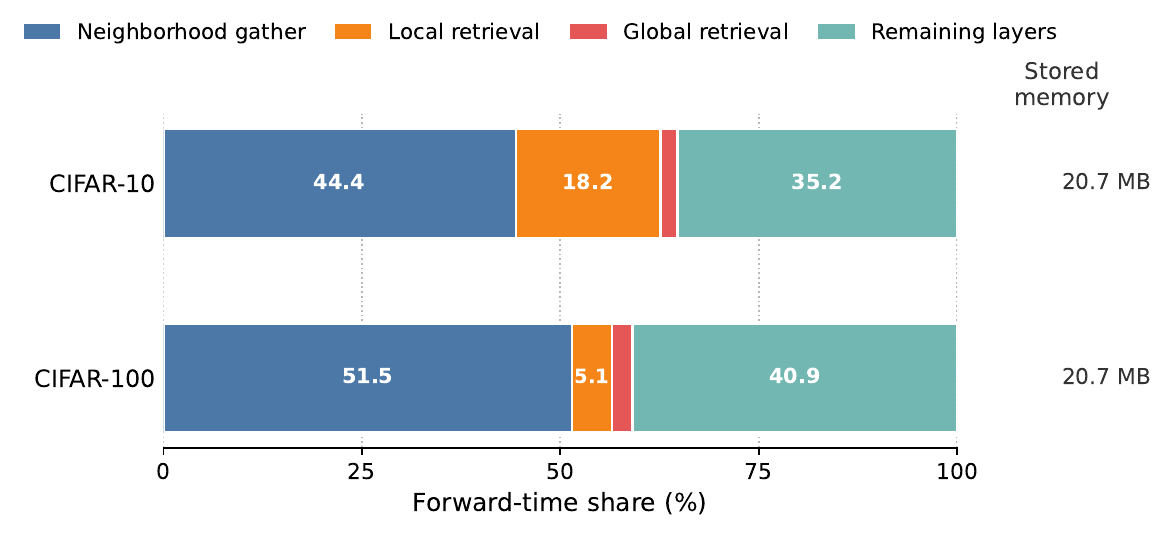}
    \caption{Runtime breakdown for V-HMN at CIFAR resolution. ``Neighborhood gather'' forms each token's local spatial neighborhood before local retrieval. ``Local retrieval'' and ``global retrieval'' are Hopfield-style memory reads from the local and global banks. ``Remaining layers'' groups patch embedding, positional encodings, projections, fusion, normalization, and classifier head. Stored memory is reported separately from trainable model parameters.}
    \label{fig:full400_vhmn_runtime_breakdown}
\end{figure}

The runtime and memory analysis quantifies the compute and memory footprint of explicit retrieval in the populated-memory 400-epoch setting. Trainable model parameters are reported separately from stored memory entries, so memory storage is not conflated with learned model weights. Table~\ref{tab:full400_complexity_runtime} compares metrics common to all listed models, while Figure~\ref{fig:full400_vhmn_runtime_breakdown} isolates where V-HMN spends forward-pass time among neighborhood gathering, local retrieval, global retrieval, and the remaining layers.

\input{generated_results/tab_full400_complexity_runtime}

Figure~\ref{fig:full400_vhmn_runtime_breakdown} shows that neighborhood gathering is the largest measured V-HMN-specific component in both datasets. Explicit local/global retrieval accounts for 20.4\% of measured CIFAR-10 forward time and 7.6\% of measured CIFAR-100 forward time, while stored memory remains about 20.7 MB and is not counted as trainable model parameters. The analysis therefore separates three quantities that are easy to conflate: trainable weights, stored latent memory entries, and the runtime cost of forming neighborhoods and reading those entries.

\FloatBarrier

\begin{table}[h]
\centering
\caption{Top-1 and top-5 retrieval hit rate, in percent (\%). Measured on CIFAR-10, CIFAR-100 and Fashion-MNIST.} 
\vspace{1ex}
\label{tab:retrieval}
\resizebox{0.65\textwidth}{!}{%
\begin{tabular}{cccc}
\toprule
 & CIFAR-10   & CIFAR-100   & Fashion-MNIST \\
\midrule
Local Top-1 Hit Rate & 30.87 & 1.92 & 35.73 \\
Local Top-5 Hit Rate & 61.38 & 6.73 & 73.44 \\
Global Top-1 Hit Rate & 36.32 & 7.82 & 68.80 \\
Global Top-5 Hit Rate & 81.43 & 24.19 & 96.24 \\

\bottomrule
\end{tabular}
}
\end{table}

\begin{table}[h]
\centering
\caption{Comparison of different retrieval methods, in terms of accuracy (\%). Measured on CIFAR-10, CIFAR-100.  Results are reported as mean $\pm$ standard deviation over 3 seeds.} 
\vspace{1ex}
\label{tab:retrieval_method}
\resizebox{0.45\textwidth}{!}{%
\begin{tabular}{ccc}
\toprule
 & CIFAR-10   & CIFAR-100    \\
\midrule
Nearest Prototype  & $93.74_{\pm 0.09}$ & $75.93_{\pm 0.16}$ \\
Top-5 Average & $93.82_{\pm 0.13}$  & $76.16_{\pm 0.12}$  \\
Hopfield Retrieval & $93.94_{\pm 0.11}$  & $76.58_{\pm 0.09}$   \\ 
\bottomrule
\end{tabular}
}
\end{table}

\subsection{Retrieval}
We analyze how the model interacts with the memories by applying it to samples from CIFAR-10, CIFAR-100 and Fashion-MNIST and calculating how frequently the memory slots that are weighted the highest in the refinement steps come from the same class as the sample input. We consider both the top-1 hit rate, which measures how frequently the single highest-weighted memory slot is from the same class as the sample input, and the top-5 hit rate, which measures how frequently at least one of five highest-weighted memory slots is from the same class as the sample input. We measure these values in the last layer of the model. The results are shown in Table~\ref{tab:retrieval}.
If the retrieval was completely random the expected top-1 and top-5 hit rates would be 1\% and 5\%, respectively, for CIFAR-100, and 10\% and 50\%, respectively, for CIFAR-10 and Fashion-MNIST. 
In all cases, the hit rate is above the random baseline, with especially clear gains for the global memory branch.
This shows that the refinement process works as expected: The latent vectors are refined towards the stored memories of the same class.

In addition, we further evaluate the contribution of the retrieval mechanism itself by comparing Hopfield retrieval with simpler alternatives under a matched-memory setting. Specifically, we keep the memory bank unchanged, including its class-balanced and label-aware structure, and only replace the retrieval function with nearest-prototype retrieval and top-$5$ averaging. 
The results are reported in Table~\ref{tab:retrieval_method}. While all methods benefit from the presence of the memory bank, Hopfield retrieval consistently achieves higher accuracy across datasets. This indicates that the performance gain cannot be solely attributed to the memory bank design or the embedded label information. Instead, the way in which the model retrieves and integrates information from memory plays a critical role. 
These findings provide empirical evidence that the proposed retrieval mechanism is not merely a redundant component, but a key factor contributing to the effectiveness of V-HMN.

\begin{figure}[h]
    \centering
    \begin{subfigure}[t]{0.45\textwidth}
        \centering
        \includegraphics[width=\textwidth]{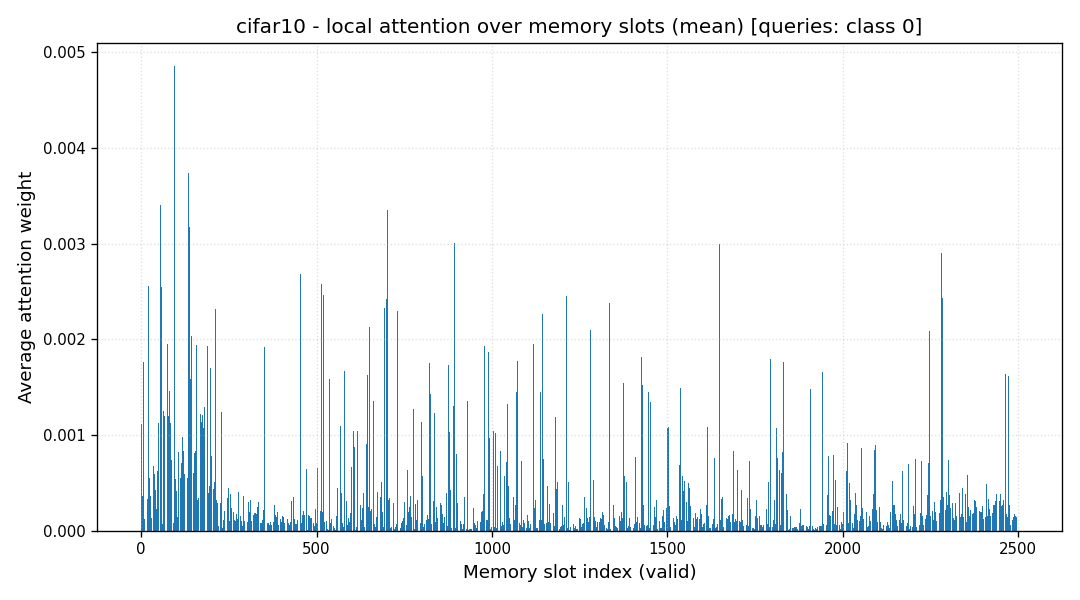}
        \caption{CIFAR-10 local memory weighting}
    \end{subfigure}
    \hfill
    \begin{subfigure}[t]{0.45\textwidth}
        \centering
        \includegraphics[width=\textwidth]{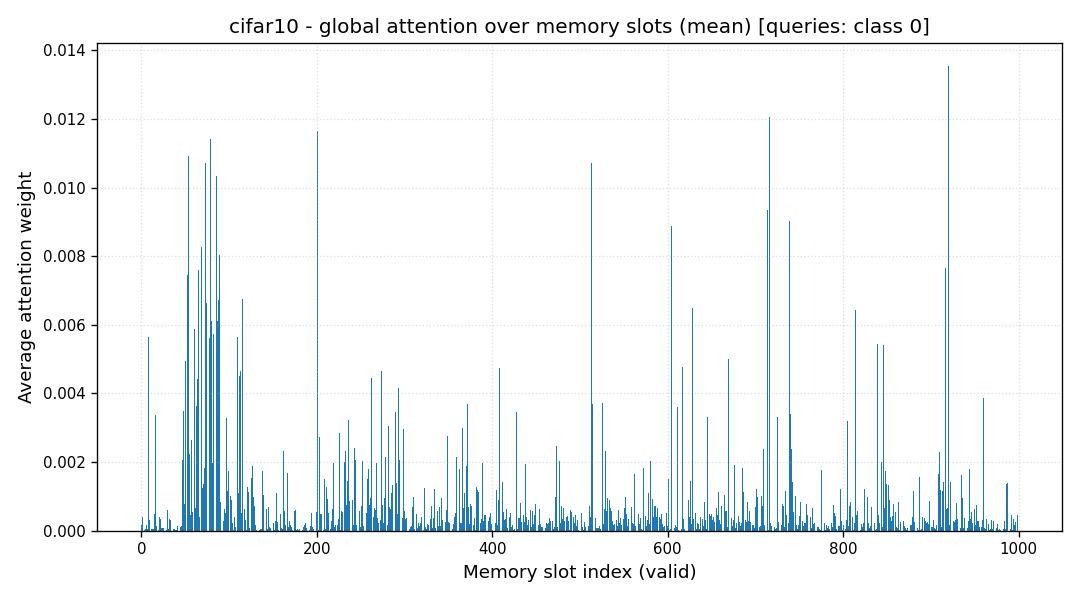}
        \caption{CIFAR-10 global memory weighting}
    \end{subfigure}
     
    \begin{subfigure}[t]{0.45\textwidth}
        \centering
        \includegraphics[width=\textwidth]{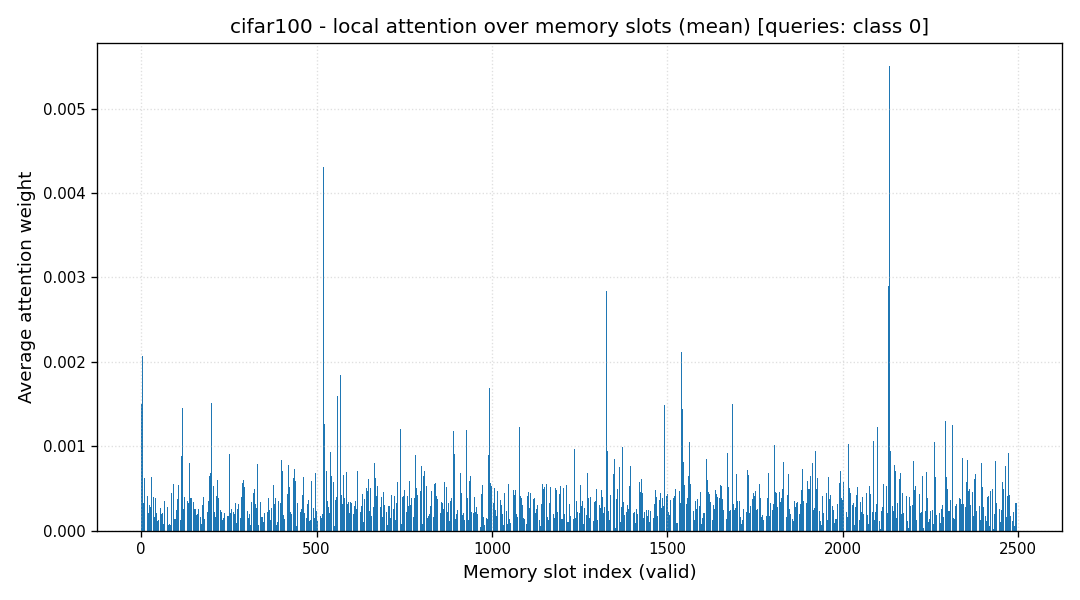}
        \caption{CIFAR-100 local memory weighting}
    \end{subfigure}
    \hfill
    \begin{subfigure}[t]{0.45\textwidth}
        \centering
        \includegraphics[width=\textwidth]{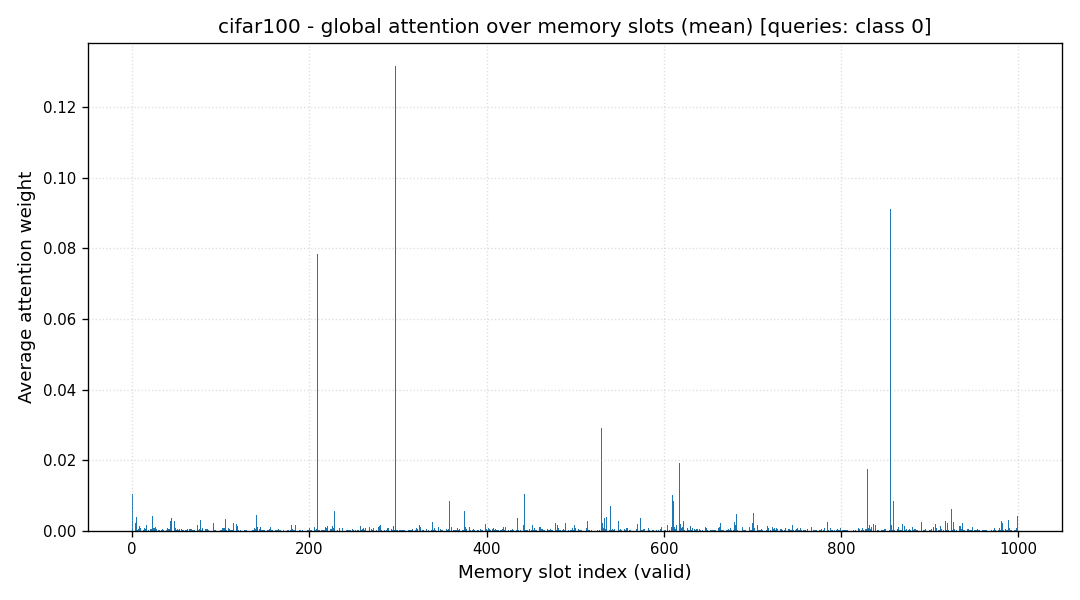}
        \caption{CIFAR-100 global memory weighting}
    \end{subfigure} 
    \begin{subfigure}[t]{0.45\textwidth}
        \centering
        \includegraphics[width=\textwidth]{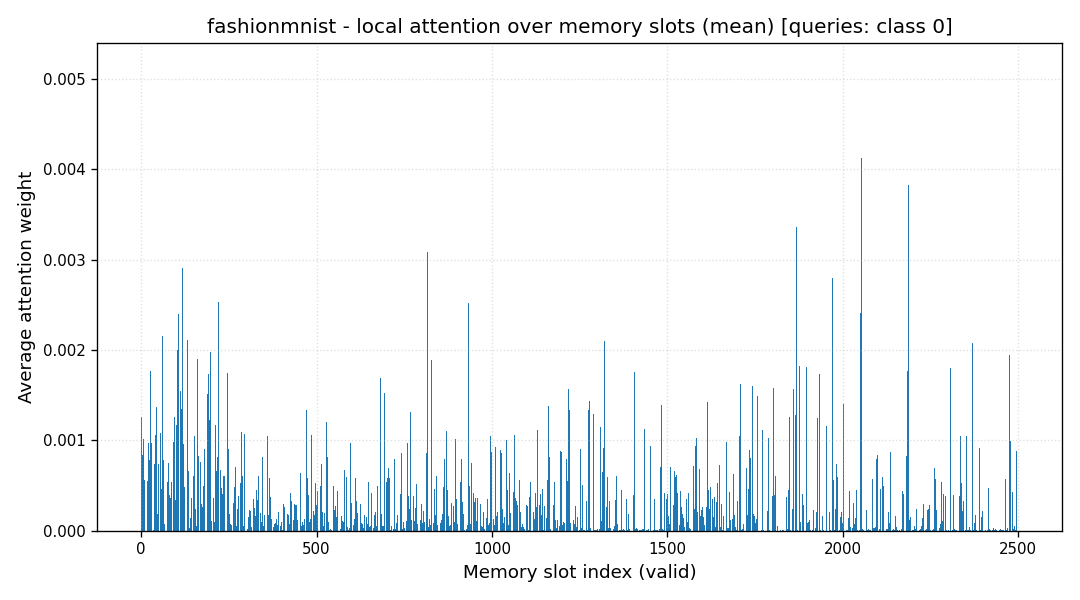}
        \caption{Fashion-MNIST local memory weighting}
    \end{subfigure}
    \hfill
    \begin{subfigure}[t]{0.45\textwidth}
        \centering
        \includegraphics[width=\textwidth]{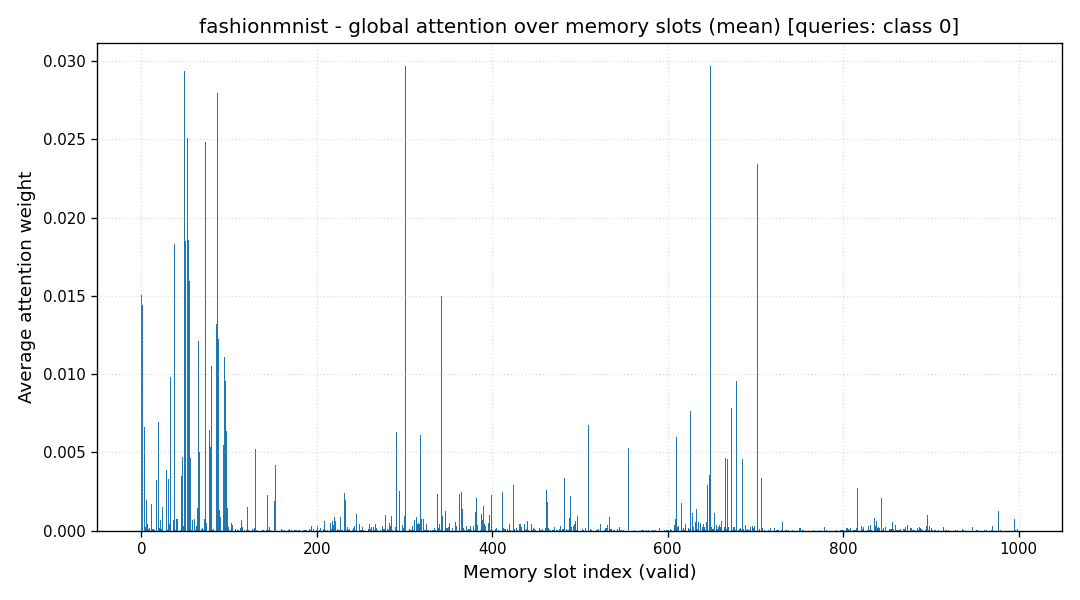}
        \caption{Fashion-MNIST global memory weighting}
    \end{subfigure}
    \caption{Visualization of the memory slot weightings of the last layer of the V-HMN}
    \label{fig:memory_weights}
\end{figure}

\subsection{Visualization of Memory Weights}

We visualize how memory slots are weighted in the last layer of the model when inputs pass through it. We perform this visualization for CIFAR-10, CIFAR-100 and Fashion-MNIST. For each, we pass all inputs from the dataset's first class through the model, and average over the resulting memory slot weightings. The results are shown in Figure~\ref{fig:memory_weights}. For CIFAR-10 and Fashion-MNIST, the weights are highest on average in the first 10\% of the memory slots, corresponding to the first dataset class. This shows that the refinement step refines the latent vectors towards prototypes of the same class. For CIFAR-100, no such patterns can be distinguished in the visualization, as the first class is only represented by the first 1\% of memory slots, which is too little to be clearly visible.



\begin{figure}[h]
    \centering
    \includegraphics[width=0.8\linewidth]{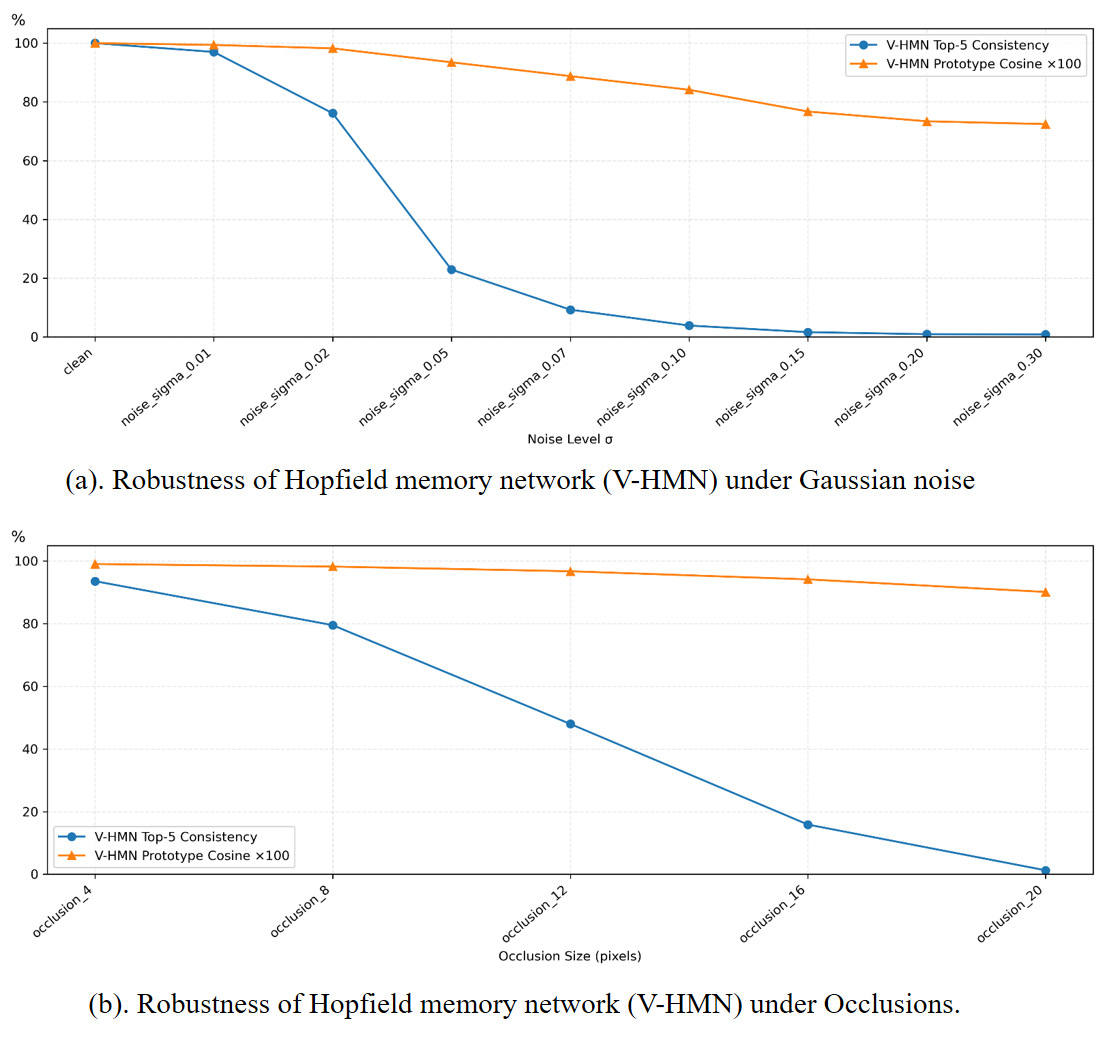}
    \caption{
    Robustness analysis of V-HMN’s associative memory under Gaussian noise and occlusion.
    }
    \label{fig:robustness2}
\end{figure}

\subsection{Associative Memory Dynamics in V-HMN: Robustness, Interpretability, and Data Efficiency}

To evaluate the robustness of V-HMN's memory module as a \emph{prototype memory}, we apply controlled perturbations to CIFAR-10 test images and examine how the retrieved prototypes change. We consider two perturbation types: (1) additive Gaussian noise with standard deviations $\sigma \in [0.01, 0.30]$, and (2) block occlusions with side lengths from $4$ to $20$ pixels (covering up to $39\%$ of the input).
To quantify how retrieval behaves under corruption, we evaluate two complementary metrics:

\begin{itemize}
\item \textbf{Top-5 Consistency (blue).}
This metric evaluates the stability of \emph{discrete} prototype indexing under perturbations.
For each clean test image, we first record the set of its top-5 retrieved prototypes.
After applying a perturbation, we obtain the corrupted image’s top-1 retrieved prototype, and check whether this prototype is contained in the clean image’s top-5 set.
We repeat this for all test images and report the percentage of cases in which the corrupted top-1 prototype remains within the clean top-5 set.
Higher values therefore indicate that the model maintains stable prototype choices even when the input is perturbed.

\item \textbf{Prototype Cosine Similarity (orange).}
This metric evaluates the \emph{semantic} stability of prototype retrieval.
For each test image, we compute the cosine similarity between:
(i) the prototype vector retrieved as top-1 for the clean image, and
(ii) the prototype vector retrieved as top-1 for its corrupted version.
We average this cosine similarity across all test samples.
High similarity indicates that even when the discrete index changes, the retrieved prototypes remain semantically similar (e.g., neighbors in prototype space).
\end{itemize}

In the occlusion experiments, the blue curve drops linearly and rapidly (falling below $20\%$ at a $16 \times 16$ occlusion), indicating frequent index switching. In contrast, the orange curve remains highly robust, maintaining over $90\%$ similarity even with $20 \times 20$ occlusions. This shows that V-HMN often switches to semantically adjacent ``neighbor'' prototypes, effectively performing pattern completion. The noise experiments exhibit a similar trend: while the blue metric is highly sensitive (dropping to $\sim20\%$ at $\sigma=0.05$), the orange metric retains around $70\%$ similarity even under extreme noise ($\sigma=0.30$), and approximately $90\%$ at $\sigma=0.07$. These results confirm that the associative retrieval mechanism provides strong denoising capabilities, mapping corrupted signals back toward the correct prototype manifold.

These findings demonstrate that V\text{-}HMN possesses strong semantic robustness, interpretability, and data efficiency. Across both noise and occlusion perturbations, the cosine similarity between pre- and post-perturbation prototypes remains remarkably stable, indicating that the model consistently maps corrupted inputs back toward nearby semantic centers in memory. This behavior reflects an effective many-to-one compression from high-dimensional pixel variations to a low-dimensional, structured prototype space, providing a principled explanation for V\text{-}HMN's data efficiency: diverse corrupted variants are absorbed into a small set of meaningful prototypes, reducing the need to observe every possible input configuration during training. Moreover, the model's ability to maintain high semantic similarity even when the discrete index changes underscores its interpretability—prototype switching occurs primarily among semantically adjacent neighbors, revealing a clear structure in the associative-memory dynamics. Finally, the smooth and predictable changes observed under increasing perturbation levels indicate stable and reliable retrieval behavior, without abrupt or erratic transitions. Together, these properties highlight the role of associative-memory refinement not merely as an architectural component, but as a robust inductive bias that enhances stability, generalization, and interpretability across challenging input conditions.

\subsection{Explorations on Spatial Window Size}

\begin{table}[h]
\centering
\caption{Ablation on spatial window size $k$. Accuracy (\%) is reported on CIFAR-10, CIFAR-100, and Fashion-MNIST. Results are reported as mean $\pm$ standard deviation over 3 seeds.}
\vspace{1ex}
\label{tab:win-size}
\resizebox{0.6\textwidth}{!}{%
\begin{tabular}{cccc}
\toprule
Window size $k$ & CIFAR-10   & CIFAR-100   & Fashion-MNIST \\
\midrule
3 & $93.94_{\pm 0.11}$  & $76.58_{\pm 0.09}$  &  $92.27_{\pm 0.06}$  \\
5 &  $93.22_{\pm 0.07}$  &  $74.80_{\pm 0.07}$     &  $91.94_{\pm 0.14}$ \\
7 & $92.60_{\pm 0.67}$  &  $70.67_{\pm 0.08}$    &  $91.23_{\pm 0.35}$\\
\bottomrule
\end{tabular}
}
\end{table}
 
Table~\ref{tab:win-size} reports the effect of varying the spatial window size $k$ for local memory retrieval. 
We observe that a smaller window ($k=3$) yields the best results across datasets, while larger windows ($k=5,7$) lead to slightly lower accuracy. 
This suggests that restricting memory matching to a compact neighborhood is beneficial, as it enforces stronger locality priors and avoids interference from irrelevant patches. 
At the same time, the overall performance difference remains small, indicating that the model is robust to the choice of $k$.

\subsection{Explorations on Size of Memory}

\begin{table*}[h]
\centering
\caption{Ablation study on the effect of local and global memory sizes in V-HMN. Top-1 test accuracy (\%) are reported as mean $\pm$ standard deviation over 3 seeds.} 
\label{tab:memory_ablation} 
\resizebox{0.85\textwidth}{!}{%
\begin{tabular}{lccc|lccc}
\toprule
\multicolumn{4}{c|}{\textbf{Local memory size}} & \multicolumn{4}{c}{\textbf{Global memory size}} \\
\midrule
Size & CIFAR-10 & CIFAR-100 & Fashion-MNIST & Size & CIFAR-10 & CIFAR-100 & Fashion-MNIST \\
\midrule
1500 &  $93.93_{\pm 0.09}$ &$ 76.47_{\pm 0.29}$ & $92.32_{\pm 0.07}$  & 500  & $93.91_{\pm 0.10}$ &$ 75.93_{\pm 0.12}$ & $92.23_{\pm 0.10}$  \\
2500 &  $93.94_{\pm 0.11}$  & $76.58_{\pm 0.09}$  &  $92.27_{\pm 0.06}$ & 1000 &  $93.94_{\pm 0.11}$  & $76.58_{\pm 0.09}$  &  $92.27_{\pm 0.06}$  \\
3500 &  $94.14_{\pm 0.01}$  & $76.42_{\pm 0.14}$  &  $92.36_{\pm 0.14}$  & 1500 &  $94.15_{\pm 0.17}$  & $76.41_{\pm 0.03}$  &  $92.12_{\pm 0.04}$  \\
4500 &  $94.04_{\pm 0.05}$  & $76.59_{\pm 0.13}$  &  $92.44_{\pm 0.03}$  & 2000 &  $93.94_{\pm 0.14}$  & $76.19_{\pm 0.04}$  &  $92.09_{\pm 0.17}$ \\
\bottomrule
\end{tabular}}
 
\end{table*}

Table~\ref{tab:memory_ablation} summarizes the effect of varying the sizes of the local and global memory banks. Performance does not grow monotonically with larger memories; instead, the best results arise from moderate capacities, approximately 2500 slots for local memory and 1000 slots for global memory. The same qualitative trend holds across all three datasets, suggesting that the effect is systematic rather than random. 
This pattern complements the iteration ablation: because the model performs only a small number of refinement steps, what matters most is having a well-curated, high-quality prototype set that can provide targeted corrections, rather than an excessively large bank.

When the memories are too small, they under-cover the feature space and limit the corrective power of each refinement step. When they are excessively large, the bank becomes redundant and introduces retrieval noise, slightly reducing accuracy and weakening the influence of each update. Together with the iteration ablation in Table~\ref{tab:iteration_ablation}, these results indicate that associative retrieval functions as a lightweight, prototype-based prior on top of a strong feedforward backbone. While not required for basic recognition (as seen from the small drop at $t=0$), the learned memories and 1–2 refinement iterations consistently improve robustness and data efficiency once an appropriate prototype set is established.

\subsection{Explorations on \texorpdfstring{$\beta$}{beta} Initialization} 
In this section, we study how the initialization of $\beta$ affects performance when the
number of refinement steps is fixed to $t=1$.

\begin{figure}[h]
    \centering
    \includegraphics[width=0.85\linewidth]{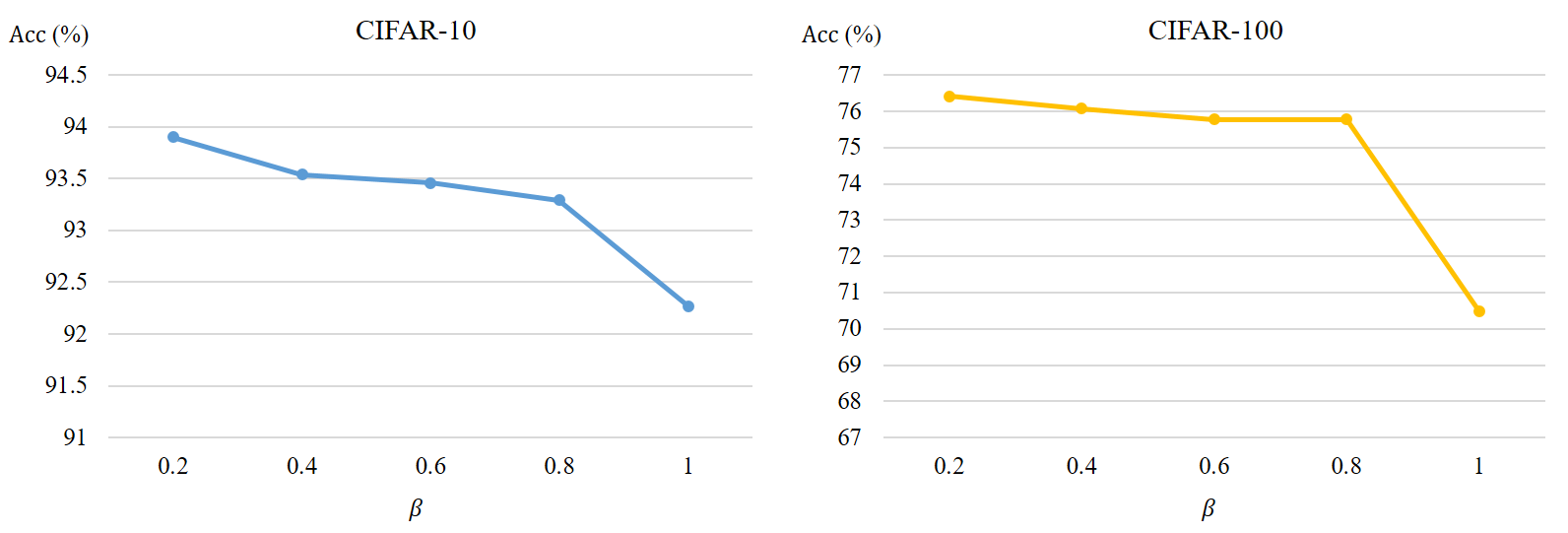}
    \caption{
    Effect of $\beta$ initialization on model accuracy with refinement iteration fixed to $t=1$. 
    }
    \label{fig:beta_init}
\end{figure}

Figure~\ref{fig:beta_init} reports results on CIFAR-10 and CIFAR-100 for
$\beta \in \{0.2, 0.4, 0.6, 0.8, 1.0\}$ (learned thereafter during training).
Performance is relatively stable for small-to-moderate values, with the best
accuracy obtained when initializing $\beta$ around $0.2$.
Setting $\beta=1.0$ consistently degrades accuracy on both datasets.

We hypothesize that the effect primarily reflects the strength of the refinement update. Very large initial values make the update overly aggressive—effectively pushing the representation too close to the retrieved prototype—while very small values under-utilize the memory-based correction.
Intermediate initialization therefore provides a balanced step size that avoids both under-correction and over-writing, which explains the better performance observed for moderate $\beta$ values. This effect is  important in the early stages of training, when the memory banks are still noisy and prototypes are less well formed; overly aggressive updates at this stage can amplify noise and hurt generalization. 

Unless otherwise noted, we initialize $\beta$ to $0.2$ (and keep it learnable),
which provides a robust starting point and yields the best or near-best results
across datasets in this single-iteration setting.

\begin{table}[h]
\centering
\caption{Performance under class-imbalance on CIFAR-10 and CIFAR-100. 
All results are top-1 accuracy (\%). Baselines include ViT \citep{dosovitskiy2020vit}, 
Swin-ViT \citep{liu2021swin}, 
MLP-Mixer \citep{tolstikhin2021mlpmixer}, 
MetaFormer \citep{yu2022metaformer}, 
Vim \citep{zhu2024visionmamba}, 
and AiT \citep{sun2025associative}. Results are reported as mean $\pm$ standard deviation over 3 seeds.}
\label{tab:imbalance_cifar}
 \vspace{1ex}
\resizebox{0.7\linewidth}{!}{%
\begin{tabular}{lcccc}
\toprule
\multirow{2}{*}{Method} &
\multicolumn{2}{c}{Imbalanced CIFAR-10} &
\multicolumn{2}{c}{Imbalanced CIFAR-100} \\
\cmidrule(lr){2-3} \cmidrule(lr){4-5}
& 50 & 100 & 50 & 100 \\
\midrule
ViT                      & $71.47_{\pm 0.34}$ &  $64.61_{\pm 0.98}$ &     $42.91_{\pm 0.29}$ & $37.45_{\pm 0.16}$ \\
Swin-ViT                 &   $65.66_{\pm 0.13}$  & $34.13_{\pm 0.15}$  & $38.37_{\pm 0.08}$  & $34.13_{\pm 0.15}$  \\
MLP-Mixer                & $72.61_{\pm 0.67}$  & $66.89_{\pm 0.14}$  & $41.90_{\pm 0.21}$ & $37.47_{\pm 0.07}$ \\
Vim             & $62.19_{\pm 0.70}$ & $57.41_{\pm 0.24}$ & $38.69_{\pm 0.12}$ & $34.55_{\pm 0.11}$ \\
AiT  & $63.47_{\pm 0.13}$ & $56.89_{\pm 0.19}$ & $38.12_{\pm 0.18}$ & $33.49_{\pm 0.22}$ \\
Meta-Former              & $55.63_{\pm 0.91}$ & $49.32_{\pm 0.23}$ & $29.47_{\pm 0.14}$ & $26.39_{\pm 0.16}$ \\
V-HMN  &  {$\boldsymbol{77.09_{\pm 0.22}}$} & {$\boldsymbol{70.43_{\pm 0.42}}$} & {$\boldsymbol{47.82_{\pm 0.11}}$} & {$\boldsymbol{42.16_{\pm 0.25}}$} \\
\bottomrule
\end{tabular}
}
\end{table}

\subsection{Imbalanced Setting}

To further evaluate the data efficiency of V-HMN, we additionally assess the model
under class-imbalanced CIFAR-10/100, following the long-tailed evaluation protocol in
prior work such as \citep{cao2019learning}, \citep{zhou2020bbn}, and \citep{kang2019decoupling}. 
We construct imbalance ratios of 50 and 100, where an imbalance ratio of~$r$ 
means that the number of samples in the most frequent class is $r$ times larger 
than that in the least frequent class. 
This setting not only makes the training data distribution highly skewed, 
but also induces imbalance in the learned prototype memories themselves, 
since minority classes contribute far fewer instances to the memory-update process.

Table~\ref{tab:imbalance_cifar} reports the results, in which 50 and 100 denote the imbalance ratios. 
Across all imbalance ratios and datasets, V-HMN achieves the strongest robustness, 
substantially outperforming ViT, Swin-ViT, MLP-Mixer, Vim, 
AiT, and Meta-Former. 
We attribute this behavior to the associative-memory mechanism: 
unlike purely feedforward models, V-HMN maintains a set of learned prototypes 
that aggregate information across the entire training distribution. 
During inference, the refinement step retrieves class-relevant prototypes 
and corrects the latent representation toward them. 
Even under severe imbalance, minority-class prototypes remain available in the memory bank 
and continue to provide stabilizing signals, mitigating representation drift 
and reducing majority-class dominance. 

In summary, these results suggest that the memory-based associative refinement 
acts as an inherent regularizer in long-tailed regimes, 
allowing V-HMN to maintain accuracy where other architectures degrade more severely.



\end{document}

%% file: generated_results/tab_full400_retrieval_disruption.tex
\begin{table}[!htbp]
\centering
\caption{400-epoch paired retrieval-disruption summary. Each row evaluates the same class-balanced checkpoints under standard retrieval and matched inference-time interventions. Each intervention cell reports Top-1 drop / changed predictions: Top-1 drop is the percentage-point decrease relative to that row's own Standard Top-1, and changed predictions is the percentage of test examples whose predicted class differs from standard retrieval. Sample memory at evaluation states whether class-indexed sample-memory entries are available for local/global reads. In Strong aug., MixUp/CutMix batches are used for the loss but are not written into class-indexed sample memory, so memory-intervention cells do not provide applicable numerical intervention measurements. Strong aug.+clean mem. uses the same strong-augmentation objective and schedule and adds a clean-label no-gradient pass after each epoch to populate sample memory.}
\label{tab:full400_retrieval_disruption}
\scriptsize
\setlength{\tabcolsep}{3.2pt}
\resizebox{\linewidth}{!}{%
\begin{tabular}{llllccccc}
\toprule
Dataset & Setting & \makecell{Sample memory\\at eval.} & Standard Top-1 & \makecell{No local\\Top-1 drop /\\changed pred.} & \makecell{No global\\Top-1 drop /\\changed pred.} & \makecell{No both\\Top-1 drop /\\changed pred.} & \makecell{Shuffled mem.\\Top-1 drop /\\changed pred.} & \makecell{Random top-$k$\\Top-1 drop /\\changed pred.} \\
\midrule
CIFAR-10 & Clean labels & populated & 91.11 $\pm$ 0.06 & 61.94/70.86 & 0.70/3.35 & 66.17/75.27 & 42.69/50.80 & 16.34/23.31 \\
CIFAR-10 & Strong aug. & not populated & 93.67 $\pm$ 0.02 & \multicolumn{5}{c}{not applicable (sample memory not populated)} \\
CIFAR-10 & Strong aug.+clean mem. & populated & 93.56 $\pm$ 0.25 & 1.24/3.95 & 0.05/0.28 & 1.23/3.98 & 14.69/19.32 & 23.71/28.84 \\
\addlinespace[1pt]
CIFAR-100 & Clean labels & populated & 65.45 $\pm$ 0.26 & 15.91/41.70 & 5.70/27.46 & 19.95/48.49 & 22.61/51.20 & 13.45/39.36 \\
CIFAR-100 & Strong aug. & not populated & 75.83 $\pm$ 0.11 & \multicolumn{5}{c}{not applicable (sample memory not populated)} \\
CIFAR-100 & Strong aug.+clean mem. & populated & 75.86 $\pm$ 0.10 & 3.63/14.47 & 0.27/5.98 & 3.74/15.15 & 11.11/26.80 & 10.74/28.15 \\
\addlinespace[1pt]
\bottomrule
\end{tabular}}
\end{table}

%% file: generated_results/tab_full400_memory_bank_properties.tex
\begin{table}[!htbp]
\centering
\caption{Memory construction and storage summary for the 400-epoch memory-bank comparison. Example-written entries are detached projected features written from labeled training examples: local entries are local-neighborhood features, and global entries are image-level mean-pooled features. These entries are latent vectors, not raw images. Gradient-learned entries are free memory vectors optimized by back-propagation. Local/global entries report the number of local and global memory entries exposed to the same Hopfield-style retrieval rule. CB-ring, U-ring, and Learned proto. use the matched 2500 local and 1000 global entries. EMA proto. stores one running class-mean prototype per class for local and global retrieval, giving 10/10 entries on CIFAR-10 and 100/100 on CIFAR-100. Stored memory is reported separately from trainable model parameters. A dash in stored memory means that no separate example-written memory buffer is used; learned prototype capacity is included in trainable parameters.}
\label{tab:full400_memory_bank_properties}
\scriptsize
\setlength{\tabcolsep}{3.2pt}
\resizebox{\linewidth}{!}{%
\begin{tabular}{lllccc}
\toprule
Design & \makecell{Example-\\written} & \makecell{Gradient-\\learned} & \makecell{Local/global\\entries} & \makecell{Stored mem.\\C10/C100 (MB)} & \makecell{Trainable params\\C10/C100 (M)} \\
\midrule
CB-ring & yes & no & 2500/1000 & 20.69/20.70 & 7.12/7.15 \\
U-ring & yes & no & 2500/1000 & 20.69/20.69 & 7.12/7.15 \\
Learned proto. & no & yes & 2500/1000 & \textemdash & 12.50/12.52 \\
EMA proto. & no & no & \makecell{10/10 (C10)\\100/100 (C100)} & 0.28/1.34 & 7.12/7.15 \\
\bottomrule
\end{tabular}}
\end{table}

%% file: generated_results/tab_full400_complexity_runtime.tex
\begin{table}[!htbp]
\centering
\caption{Runtime and compute comparison for the populated-memory 400-epoch setting. Approx. FLOPs are profiler-derived per-image floating-point operations. Throughput and GPU memory are measured consistently across the listed models. V-HMN stored-memory and retrieval-time components are separated in Figure~\ref{fig:full400_vhmn_runtime_breakdown} and the surrounding text.}
\label{tab:full400_complexity_runtime}
\scriptsize
\setlength{\tabcolsep}{3.2pt}
\resizebox{\linewidth}{!}{%
\begin{tabular}{llcccc}
\toprule
Dataset & Model & Trainable params (M) & Approx. FLOPs (G) & Throughput (img/s) & GPU mem. (GB) \\
\midrule
CIFAR-10 & V-HMN & 7.12 & 1.85 & 8874 & 1.08 \\
CIFAR-10 & ViT & 7.16 & 0.92 & 29360 & 0.38 \\
CIFAR-10 & MLP-Mixer & 8.71 & 1.35 & 19808 & 0.67 \\
\addlinespace[1pt]
CIFAR-100 & V-HMN & 7.15 & 0.96 & 10692 & 0.72 \\
CIFAR-100 & ViT & 7.19 & 0.92 & 29114 & 0.38 \\
CIFAR-100 & MLP-Mixer & 8.76 & 1.35 & 19943 & 0.67 \\
\addlinespace[1pt]
\bottomrule
\end{tabular}}
\end{table}